\documentclass[preprint,12pt]{elsarticle}

\usepackage{amssymb}
\usepackage{amsmath,amssymb,amsfonts}
\usepackage{array}
\usepackage{algorithmic}
\usepackage{graphicx}
\usepackage{textcomp}
\usepackage{caption}
\usepackage{subcaption}
\usepackage{xcolor}
\usepackage{bm}
\usepackage{soul}
\usepackage{booktabs}
\usepackage{tikz} 
\usepackage[colorlinks=true, allcolors=blue]{hyperref}
\usepackage{epstopdf}
\usepackage{cleveref}
\usepackage{makecell}
\usepackage{duckuments}
\usepackage{comment}
\usepackage[normalem]{ulem}
\usepackage{comment}
\usepackage{xcolor}
\usepackage{booktabs}
\usepackage{graphicx}
\usepackage{float}
\definecolor{mydarkred}{RGB}{139,0,0} 
\usetikzlibrary{shapes.geometric, arrows}
\usepackage{xcolor}

\journal{Robots and Autonumous Systems}

\begin{document}

\begin{frontmatter}
\title{FG-PE: Factor-graph Approach for Multi-robot Pursuit-Evasion}

\author{Messiah Abolfazli Esfahani\corref{cor1}}
\ead{mabolfazli@torontomu.ca}
\address{Toronto Metropolitan University, Canada}

\author{Ayşe Başar}
\ead{ayse.bener@torontomu.ca}
\address{Toronto Metropolitan University, Canada}

\author{Sajad Saeedi}
\ead{s.saeedi@torontomu.ca}

\address{Toronto Metropolitan University, Canada}
\cortext[cor1]{Corresponding author}

\begin{abstract}
With the increasing use of robots in daily life, there is a growing need to provide robust collaboration protocols for robots to tackle more complicated and dynamic problems effectively. This paper presents a novel, factor graph-based approach to address the pursuit-evasion problem, enabling accurate estimation, planning, and tracking of an evader by multiple pursuers working together. It is assumed that there are multiple pursuers and only one evader in this scenario. The proposed method significantly improves the accuracy of evader estimation and tracking, allowing pursuers to capture the evader in the shortest possible time and distance compared to existing techniques. In addition to these primary objectives, the proposed approach effectively minimizes uncertainty while remaining robust, even when communication issues lead to some messages being dropped or lost. Through a series of comprehensive experiments, this paper demonstrates that the proposed algorithm consistently outperforms traditional pursuit-evasion methods across several key performance metrics, such as the time required to capture the evader and the average distance traveled by the pursuers. Additionally, the proposed method is tested in real-world hardware experiments, further validating its effectiveness and applicability.
\end{abstract}

\end{frontmatter}


\section{Introduction}
\label{sec:sample1}
With the increasing presence of robots across various industries, there has been a growing emphasis on inter-robot collaboration. Collaborative robots assist with tasks such as search and rescue~\cite{queralta2020collaborative}, trajectory planning~\cite{Patwardhan2023gbpplanner}, agriculture~\cite{lytridis2021overview}, object transportation~\cite{jaafar2023mrcap}, and collision avoidance~\cite{alonso2018cooperative}. A key challenge in search and rescue operations is the pursuit-evasion (PE) problem, where pursuers try to catch evaders.

Several methods have been utilized to solve the PE problem. Classical methods use graph-based techniques, treating the PE problem as a search problem~\cite{kolling2009pursuit,ramaithitima2016hierarchical}. Reinforcement learning (RL) approaches~\cite{kartal2021optimal,zhou2022decentralized} and game theory methods~\cite{ho1965differential} have also been investigated. However,  existing methods often lack an efficient representation of the relationships between pursuers, evaders, and their environment. For instance, if one pursuer loses its history of movements, retrieving that information can be challenging. Utilizing a graph-based approach can establish structural connections between entities, improving the overall understanding of the situation. In addition, the aforementioned methods do not account for uncertainties in predictions and do not facilitate message passing among robots. This limitation affects the scalability and versatility of PE variants.

In this paper, we present FG-PE, a method that employs factor graphs (FG)~\cite{factor-graphs-for-perception} to tackle the PE problem. Factor graphs provide a flexible framework capable of adapting to varying numbers of pursuers and obstacles. This approach addresses prediction uncertainty by guiding pursuers to minimize capture time, distance, and uncertainty, utilizing information from sensor measurements and past observations. 

\begingroup\color{black}{
The main contribution of the proposed method, FG-PE, lies in formulating a factor-graph-based solution to the pursuit–evasion problem. To the best of our knowledge, this is the first work to model the PE problem using a factor graph. FG-PE achieves up to a 100-fold reduction in capture time compared to benchmark methods, while requiring fewer measurements. This leads to lower memory usage and computational overhead, enabling FG-PE to outperform Constant Bearing (CB), Pure Pursuit (PP) ~\cite{makkapati2019optimal}, and GUT~\cite{yang2022game}. The key features of FG-PE are as follows:
}\endgroup
         {(1)} FG-PE can estimate the evader's position and plan the pursuers while considering short time windows. This means that at each time step, it can determine the planned position for each pursuer for the subsequent time step.
     {(2)} FG-PE accounts for uncertainty in predictions using the probabilistic nature of factor graphs. It aims to minimize this uncertainty by providing a plan through factor graphs, which enable efficient representation and optimization in the presence of noise. Videos of the experiments are available on the website of the project:~\href{https://sites.google.com/view/pursuit-evasion}{https://sites.google.com/view/pursuit-evasion}. 
     {(3)} \textcolor{black}{FG-PE} is easily extendable to multiple pursuers and allows for high obstacle-wise scalability.
\textcolor{black}{(4) FG-PE is robust against lost messages and performs well even when some messages are dropped within the graph structure.}


\section{Literature Review and Background}
\label{sec:bkgnd}
The PE problem finds applications in various robotics scenarios, such as surveillance, search and rescue operations, and multiplayer games~\cite{kshirsagar2018survey}. PE involves two groups of agents: pursuers and evaders. The evaders aim to reach a predefined goal, such as a specific location, object, or target area, while the pursuers try to prevent them from doing so. Key aspects of the PE problem include multi-agent tracking~\cite{hu2022multi,styles2020multi} and path planning~\cite{Patwardhan2023gbpplanner}. There are three main approaches solving the PE problem: search-based, learning-based, and graph-based approaches.

A direct method for addressing the PE problem is to use search algorithms like Depth-First Search (DFS)~\cite{tarjan1972depth} or Breadth-First Search (BFS)~\cite{bundy1984breadth}.
A critical aspect of the pursuit-evasion problem is the speed of the agents. Furhan \textit{et al.}~\cite{yan2019multiagent} introduced a method that accounts for uncertainties in the pursuers' speed, which can lead to issues such as moving to unintended locations. \textcolor{black}{A cooperative strategy has been developed for multiple pursuers to capture a high-speed evader in holonomic systems, addressing both formation and capture requirements~\cite{makkapati2016cooperative}. Multi-pursuer pursuit–evasion games with a high-speed evader are analyzed using perfectly encircled formations and Apollonius circles, highlighting conditions for guaranteed capture as well as evader escape strategies~\cite{ramana2015cooperative}.} 

Learning-based algorithms are also suitable for solving the PE problem. 
Yu \textit{et al.}~\cite{yu2020distributed} proposed a method focusing on the application of reinforcement learning (RL) techniques to address PE. 
The authors propose a fully decentralized multi-agent deep RL approach. In this approach, each agent is modeled as an individual deep RL agent with its learning system, including an individual action-value function, learning update process, and action output. To facilitate coordination among agents, limited information about other environmental agents is incorporated into the learning process. Engin \textit{et al.} \cite{engin2021learning} aimed to solve a type of PE in which agents have limited visibility. To overcome the challenges associated with large state spaces, a new learning-based method is introduced that compresses the game state and uses it to plan actions for the players \cite{engin2021learning}. Nikolaos-Marios \textit{et. al} formulated the problem of PE as a zero-sum differential game and they proposed an online RL approach to solve it \cite{kuck2020belief}. One of the other proposed approaches for solving the PE problem is utilizing the Q-learning approach\cite{selvakumar2022min}.
\sloppy

Graph-based methods have also been utilized to solve the PE problem. 
Kirousis~\textit{et al.}~\cite{kirousis1986searching} converted the PE problem into a graph searching problem. 
Another approach to solving the problem in a graph searching manner, involves vertex pursuit, where the pursuer aims to occupy the same vertex as the evader~\cite{ramana2017pursuit}. Another type is the edge pursuit, in which the entities move along the edges of the graph, and the pursuer attempts to move to the edge occupied by the evader~\cite{von2018pursuit}.

\begingroup\color{black}{
Several studies have derived optimal evasion strategies under specific pursuer behaviors. Makkapati \textit{et al.}~\cite{Makkapati2018JGDC, makkapati2018pursuit} analyzed a two-pursuer, one-evader scenario and derived analytical strategies when the pursuers follow either constant-bearing or pure-pursuit trajectories. In a more general setting, Makkapati and Tsiotras~\cite{makkapati2019optimal} extended these results to multi-player pursuit–evasion problems by combining optimal evasion with task allocation among pursuers. These works provide important benchmarks for evaluating evader strategies, although they rely on the assumption of specific pursuer behaviors.}
\endgroup

\textcolor{black}{
Several approaches have addressed the pursuit–evasion problem in the presence of external flow fields. One method approximates the flow using time-invariant affine functions~\cite{sun2015pursuit}. Another line of work handles dynamic disturbances by defining reachable set inclusions, where level-set equations are solved to generate the pursuer’s reachable sets~\cite{sun2017pursuit}. Reachable set–based formulations have also been extended to pursuit–evasion scenarios involving one evader and multiple pursuers~\cite{sun2017multiple}. Multi-pursuer, multi-evader differential games in three-dimensional environments with dynamic disturbances have been studied using partitioning strategies and reachability analysis, enabling the computation of time-optimal trajectories and strategies~\cite{Sun2019}. More recently, single-pursuer, single-evader games in stochastic flow fields have been analyzed using forward reachability and chance-constrained linear–quadratic formulations, solved via Gauss–Seidel iterations~\cite{makkapati2022reachability}.
}

There are several real-world applications that require the collaboration of multiple robots such as object transportation \cite{an2023multi} and task allocation \cite{martin2023multi}. 
One approach to coordinating the necessary communication among robots is by utilizing graph representation. Factor graphs~\cite{factor-graphs-for-perception}, described in Sec.~\ref{sec:fg} in detail, can represent complex relationships between agents concisely and are suitable for addressing a wide range of problems such as kinodynamic motion planning that fully considers whole-body dynamics and contacts~\cite{xie2020factor}. Factor graphs have been widely used to solve various multi-robot problems, such as localization~\cite{murai2023robot}, path planning~\cite{patwardhan2023distributed}, and control~\cite{jaafar2023mrcap}.
\begingroup\color{black}{
\subsection{Factors in Factor Graphs}
Factor graphs are bipartite graphs that represent how a global function can be decomposed into a product of local functions, called factors~\cite{kschischang2002factor}.
Normally, given a set of variables $X = \{x_1, x_2, \dots, x_n\}$, a factor graph expresses the joint distribution as

\begin{equation}
f(X) = \prod_{a \in \mathcal{F}} f_a(X_a),
\end{equation}

where $\mathcal{F}$ denotes the set of factors in the factor graph, and each factor $f_a$ is a local function depending only on a subset of variables $X_a \subseteq X$.
 Factor nodes are connected to variable nodes when the corresponding variable appears in the factor. Several types of factors appear frequently in robotics and estimation problems:

\subsubsection{Unary factors} 
Unary factors depend on a single variable. They typically encode prior information or measurements about an individual variable:
\begin{equation}
f(x_i) \propto \exp\Bigg(-\frac{1}{2} \| h(x_i) - z \|^2_{\Omega} \Bigg), i=1,2,...,n
\end{equation}
where $z$ is the measurement, $h(\cdot)$ is the measurement model.  
Here $||.||$ denotes the Mahalanobis distance \cite{mclachlan1999mahalanobis}, 
and $\Omega$ represents the information matrices of the variables.
For the Gaussian distribution, $\Omega = \Sigma^{-1}$, $\Sigma$
is the covariance of the variable.
A prior factor is a special case of a unary factor. It encodes prior knowledge about a variable and effectively anchors it to a known or assumed value while allowing for uncertainty. Mathematically, a prior factor constrains a variable $x$ to a prior mean $\mu$ with information matrix $\omega$ (covariance $\Sigma=\Omega^{-1}$), and is typically modeled as a Gaussian:
\begin{equation}
f_{\text{prior}}(x) \propto \exp\Big(-\frac{1}{2} \|x - \mu\|_{\Omega}^{2}\Big),
\end{equation}
where
\begin{equation}
\|x - \mu\|_{}^{2} = (x - \mu)^\top \Omega (x - \mu)
\end{equation}
is the Mahalanobis distance between the variable $x$ and the prior mean $\mu$, weighted by the information matrices $\Omega$.

\subsection{Pairwise factors} 
Pairwise factors involve two variables and are often used to express relative constraints between them:
\begin{equation}
f(x_i, x_j) \propto \exp\Bigg(-\frac{1}{2} \| h(x_i, x_j) - z \|^2_{\Omega} \Bigg).
\end{equation}
These appear in odometry, landmark observations, or inter-robot measurements.

\subsubsection{Higher-order factors}
Higher-order factors involve more than two variables. They are used when the constraint naturally depends on a set of variables simultaneously, such as synchronization:
\begin{equation}
f(x_i, x_j, x_k) \propto \exp\Bigg(-\frac{1}{2} \| h(x_i, x_j, x_k) - z \|^2_{\Omega} \Bigg).
\end{equation}
}\endgroup

In this paper, we utilize factor graphs to address the PE problem. To the best of the authors' knowledge, there is no existing work, that formulates the PE problem using a factor graph representation.

\section{Proposed Method}\label{sec:method}
In this section, the problem statement and the proposed solution are presented. Table~\ref{table:variables} lists the notations used; 
\begingroup\color{black}{$q$ is used for the evader, and $p_j$ is used for the $j^{\text{th}}$ pursuer. Factors are denoted by $f$, with superscripts indicating the type of constraint, such as dynamics (i.e. $f^d$), measurement (i.e. $f^m$), planning (i.e. $f^{mov}$), collision avoidance (i.e. $f^c$), and obstacle avoidance (i.e. $f^o$). Each factor $f$ corresponds to an error term $e$, 
which represents the cost minimized during optimization. The same notation is used for the cost. 
For factors and costs, after the constraint type, further symbols indicate if the factor/cost is associated with a pursuer, an evader, or an obstacle. The indices $i$ and $j$ denote the identifiers of the corresponding pursuers or obstacles, while the subscript $t$ represents the discrete time step. Note, there is only one evader.
}
\endgroup
\begin{table}[h]
\centering
\small
\color{black}
\begin{tabular}{|p{1.2cm}|p{11.5cm}|}
\hline
{\footnotesize Symbol} & {\footnotesize Description} \\
\hline
\multicolumn{2}{|c|}{\textbf{Variables}} \\
\hline
\( x_{t}^q \) & {\footnotesize Evader pose in SE(2) at time step \(t\)} \\
\( \hat{x}_{t}^q \) & {\footnotesize Estimated evader pose in SE(2) at time step \(t\)} \\
\( x_{t}^p \) & {\footnotesize Pursuer pose in SE(2) at time step \(t\)} \\
$\Omega$ & {\footnotesize Information matrix of variables} \\
\hline
\multicolumn{2}{|c|}{\textbf{Factors}} \\
\hline
$f^{dq}_{t}$ & {\footnotesize Dynamic factor of the evader at time step \(t\)} \\
$f^{dp_{j}}_{t}$ & {\footnotesize Dynamic factor of the \(j^{th}\) pursuer at time step \(t\)} \\
$f^{mp_{j}}_{t}$ & {\footnotesize Measurement factor between \(j^{th}\) pursuer and the evader at time \(t\)} \\
$f^{mo_{ji}}_{t}$ & {\footnotesize Measurement factor between \(j^{th}\) pursuer and \(i^{th}\) obstacle at time \(t\)} \\
$f^{movp_{j}}_{t}$ & {\footnotesize Planning factor of the \(j^{th}\) pursuer at time \(t\)} \\
$f^{cp_{ji}}_{t}$ & {\footnotesize Collision avoidance factor between pursuer $j$  and pursuer $i$ at time \(t\)} \\
$f^{op_{ij}}_{t}$ & {\footnotesize Obstacle avoidance factor between \(j^{th}\) pursuer and \(i^{th}\) obstacle at time \(t\)} \\
\hline
\multicolumn{2}{|c|}{\textbf{Costs (Error Terms)}} \\
\hline
$e^{dq}_{t}$ & {\footnotesize Cost of dynamic factor of the evader} \\
$e^{dp_{j}}_{t}$ & {\footnotesize Cost of dynamic factor of the \(j^{th}\) pursuer} \\
$e^{mp_{j}}_{t}$ & {\footnotesize Cost of measurement factor between \(j^{th}\) pursuer and evader} \\
$e^{mo_{ji}}_{t}$ & {\footnotesize Cost of measurement factor between \(j^{th}\) pursuer and \(i^{th}\) obstacle} \\
$e^{movp_j}_{t}$ & {\footnotesize Cost of planning factor for pursuer \(j\)} \\
$e^{cp_{ji}}_{t}$ & {\footnotesize Cost of collision avoidance between pursuer $j$ and pursuer $i$} \\
$e^{op_{ji}}_{t}$ & {\footnotesize Cost of obstacle avoidance factor between pursuer \(j\) and obstacle \(i\)} \\
\hline
\end{tabular}
\color{black}
\vspace{-3 mm}
\caption{\textcolor{black}{Summary of notation. Variables represent unknown quantities to be estimated, factors encode probabilistic or planning constraints, and costs correspond to the residual errors associated with each factor.}}
\label{table:variables}
\vspace{-12pt} 
\end{table}

\subsection{Problem Statement}
In this work, we assume that a team of $N_{p}$ pursuers, indexed by \mbox{$N = {1, 2, \ldots, N_{p}}$}, and one evader ($N_{q}=1$) are exploring the $\mathbb{R}^2$ space. The linear and angular velocity of robots are bounded based on the following conditions, where $v^{max}$ and $\omega^{max}$ represent the maximum allowable linear and angular velocity for the robots:
\begin{align}
\mathcal{U} = \{ v \in \mathbb{R}^2 \mid ||v|| \leq v^{\text{max}} \}
\label{pursuer_speed_constraint}
\end{align}
\begin{align}
\mathcal{R} = \{ \omega \in \mathbb{R}^2 \mid ||\omega|| \leq \omega^{\text{max}} \}
\label{evader_speed_constraint}
\end{align}

The goal of the pursuers is to capture the evader within a defined capturing radius \( r > 0 \). The pursuers win if the following condition is satisfied:
\begin{align}
\exists  i \in N_{p}, ||x^{{p}_{i}}_{t}-x^{q}_{t}||\leq r
\label{capturing_condition}
\end{align}

In Eq.~\eqref{capturing_condition}, if at time step \( t \) the distance between a pursuer and the evader is less than the capturing radius, it indicates that the pursuers win the game. 
In this paper, we assume:
1) pursuers can observe the evaders but the measurements (range/bearing) are noisy,  
2) there are random number of obstacles in the environment that is shown by $N_o$,
3) the evader has a pre-defined goal point to reach, and
4) pursuers neither know the goal of the evader nor the motion model of the evader. 
The objective for pursuers is to 
1) estimate the position of the evader,
2) minimize the uncertainty of the estimate, and
3) capture the evader. 
\subsection{Path Planning of Evader}
The evader aims to reach a predefined goal, known only to itself. 
\textcolor{black}{The evader employs the Dynamic Window Approach (DWA), a popular method in robotics for motion planning and obstacle avoidance \cite{fox1997dynamic}. DWA is effective in scenarios where a robot navigates through moving obstacles. It dynamically adjusts the robot's velocity and steering commands based on real-time sensory input and desired goal states. By continuously evaluating and updating potential future trajectories within a dynamically defined window, the evader can efficiently navigate toward its goal while avoiding obstacles and pursuers. The evader iterates through all possible trajectories based on angular velocity and linear velocity, selecting the best movement for the next time step. In this paper, we utilize a modification of the DWA algorithm\footnote{https://github.com/RajPShinde/Dynamic-Window-Approach} for planning the \begingroup\color{black}{evader's}\endgroup\ trajectory.}
\subsection{Factor Graphs}\label{sec:fg}
This paper proposes a factor graph based solution to determine the optimal motion of pursuers. In factor graphs, nodes represent variables, such as a robot's pose, while edges represent relationships between these variables, including kinematic constraints and measurements.

\begin{figure}[t!]
\centering
\begin{tikzpicture}[scale=1]
    \foreach \pos/\name/\color in {{(0,0)/x_{0}^q/red}, {(2,0)/x_{1}^q/red},{(4,0)/x_{2}^q/red},{(6,0)/x_{3}^q/red},{(0,3)/x_{0}^{p_{1}}/green},{(2,3)/x_{1}^{p_{1}}/green},{(4,3)/x_{2}^{p_{1}}/green},{(6,3)/x_{3}^{p_{1}}/green},{(0,-3)/x_{0}^{p_{2}}/green},{(2,-3)/x_{1}^{p_{2}}/green},{(4,-3)/x_{2}^{p_{2}}/green},{(6,-3)/x_{3}^{p_{2}}/green}}
\node[circle,draw,fill=\color,minimum size=.25cm,inner sep=1pt,font=\tiny] (\name) at \pos {$\name$}; %
    \foreach \pos/\name in {{(1,0)/1},{(3,0)/3},{(5,0)/5},{(0,1.5)/f3},{(6,1.5)/f6},{(0,-1.5)/f7},{(6,-1.5)/f10},{(1,3)/f11},{(1,3)/f12},{(3,3)/f13},{(5,3)/f14}}
        \node[rectangle,draw,fill=black,minimum size=0.1cm] (\name) at \pos {}; 
    \foreach \pos/\name in {{(3,-3)/f16},{(5,-3)/f17}}
        \node[rectangle,draw,fill=black,minimum size=0.05cm] (\name) at \pos {}; 
    \foreach \pos/\name/\label/\text in {{(1,-3)/f15/$f_1^{dp_{2}}$},{(3,-3)/f16/$f_2^{dp_{2}}$},{(5,-3)/f17/$f_3^{dp_{2}}$},{(1,3)/f12/$f_1^{dp_{1}}$},{(3,3)/f13/$f_2^{dp_{1}}$},{(5,3)/f14/$f_3^{dp_{1}}$},{(1,0)/1/$f_0^{dq}$},{(3,0)/3/$f_1^{dq}$},{(5,0)/5/$f_2^{dq}$}}
        \node[rectangle,draw,fill=black,minimum size=0.1cm,label={[font=\tiny]90:\text}] (\name) at \pos {};
    \foreach \pos/\name/\label/\text in {{(0,4.5)/fo1/$f_0^{mo_{11}}$},{(2,4.5)/fo2/$f_1^{mo_{11}}$},{(4,4.5)/fo3/$f_2^{mo_{11}}$},{(6,4.5)/fo4/$f_3^{mo_{11}}$}}
        \node[rectangle,draw,fill=black,minimum size=0.1cm,label={[font=\tiny]90:\text}] (\name) at \pos {}; 
    \foreach \pos/\name/\label/\text in {{(-1,-3)/fo5/$f_0^{{mo_{21}}}$},{(7,-3)/fo8/$f_3^{{mo_{21}}}$},{(4,-1.5)/fo7/$f_2^{mo_{21}}$},{(2,-1.5)/fo6/$f_1^{mo_{21}}$}}
        \node[rectangle,draw,fill=black,minimum size=0.1cm,label={[font=\tiny]90:\text}] (\name) at \pos {}; 
    \foreach \pos/\name/\label/\text in {{(0,1.5)/$f_{0}^{mp_{1}}$},{(6,1.5)/$f_{3}^{mp_{1}}$},{(6,-1.5)/$f_{3}^{mp_{2}}$},{(0,-1.5)/$f_{0}^{mp_{2}}$}}
        \node[rectangle,draw,fill=black,minimum size=0.1cm,label={[font=\tiny]180:\text}] (\name) at \pos {};
    \foreach \source/\target/\color in {x_{0}^{p_{1}}/f11/black,f11/x_{1}^{p_{1}}/black,x_{1}^{p_{1}}/f13/black,f13/x_{2}^{p_{1}}/black,x_{2}^{p_{1}}/f14/black,f14/x_{3}^{p_{1}}/black}
        \path (\source) edge[draw=\color] (\target);
    \foreach \source/\target/\color in {x_{0}^{p_{2}}/f15/black,f15/x_{1}^{p_{2}}/black,x_{1}^{p_{2}}/f16/black,f16/x_{2}^{p_{2}}/black,x_{2}^{p_{2}}/f17/black,f17/x_{3}^{p_{2}}/black}
        \path (\source) edge[draw=\color] (\target);
    \foreach \source/\target/\color in {x_{0}^q/1/black, x_{1}^q/1/black, x_{1}^q/x_{2}^q/black, x_{2}^q/x_{3}^q/black}
        \path (\source) edge[draw=\color] (\target);
    \foreach \source/\target/\color in {fp1/x_o/black}
            \foreach \source/\target/\color in {x_{0}^{p_{1}}/f3/black,f3/x_{0}^q/black,f6/x_{3}^q/black,x_{3}^{p_{1}}/f6/black,
            x_{0}^{p_{2}}/f7/black,f7/x_{0}^q/black,x_{3}^{p_{2}}/f10/black,f10/x_{3}^q/black}
        \path (\source) edge[draw=\color,dashed] (\target);  
        \foreach \source/\target/\color in 
        {{fo1/x_{0}^{p_{1}}/black},{fo2/x_{1}^{p_{1}}/black},{fo3/x_{2}^{p_{1}}/black},{fo4/x_{3}^{p_{1}}/black},{fo5/x_{0}^{p_{2}}/black},{fo6/x_{1}^{p_{2}}/black},{fo7/x_{2}^{p_{2}}/black},{fo8/x_{3}^{p_{2}}/black}}
        \path (\source) edge[draw=\color,dotted] (\target);  
\end{tikzpicture}
\caption{The factor graph representation for the PE problem. Green nodes represent the variables associated with the poses of the pursuers. Red nodes are poses of evaders. Squares represent the constraints between the variables, including motion models (e.g. $f_1^{dp_1}$), measurements of the robots (e.g. $f_0^{mp_1}$) or landmarks (e.g. $\textcolor{black}{f_1^{mo_{11}}}$).}
\label{pursuers_moving}
\vspace{-14pt} 
\end{figure}
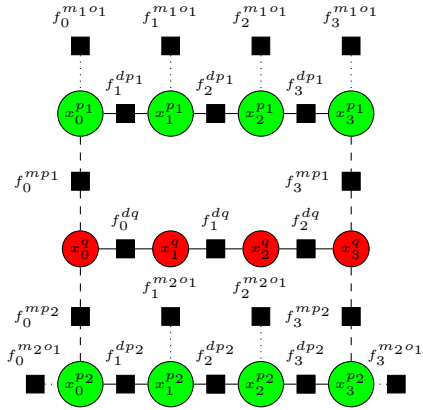

An optimization cost is defined based on minimizing the costs associated with the edges of the factor graph. \begingroup\color{black}{The optimization problem is formulated as a maximum a posteriori (MAP)~\cite{factor-graphs-for-perception} inference problem and is implemented using the Georgia Tech Smoothing and Mapping (GTSAM) library}\endgroup~\cite{dellaert2012factor}. \begingroup \color{black}{It represents inference problems using factor graphs and Bayesian networks, which provide a structured framework for optimization. These problems are then solved efficiently using sparse matrix factorization techniques to compute an optimal solution.} \endgroup  \begingroup \color{black}{The Levenberg–Marquardt (LM) algorithm}\endgroup~\cite{marquardt1963algorithm} \begingroup \color{black}{is employed as the optimizer, providing fast convergence. LM is a widely used iterative method for solving nonlinear least-squares problems, combining the advantages of the Gauss–Newton algorithm  and gradient descent.}\endgroup \ The factor graph representation of the PE problem \textcolor{black}{is shown in Fig.~\ref{pursuers_moving} and} it is adaptable and can be designed to enable the pursuers to move within the search space to actively pursue the evader. 
The overall cost function \begingroup\color{black}at time $T$ \endgroup can be defined as
\begin{align}
    \textcolor{black}{\mathcal{J}}(x^{q}_{0:T},x^{p}_{0:T}) &= \sum_{t=0}^{T}\Big( e_{t}^{\textcolor{black}{dq}} + \sum_{j=1}^{N_P} e_{t}^{mp_{j}} \nonumber 
    \quad +\sum_{j=1}^{N_p}\sum_{\substack{i=1 \\ i \neq j}}^{N_P} e_t^{\textcolor{black}{cp_{ji }}}\\&+\sum_{j=1}^{N_p}\sum_{i=1}^{N_o}\textcolor{black}{(}e_t^{\textcolor{black}{op_{ji}}}+e^{\textcolor{black}{mo_{ji}}}_t\textcolor{black}{)}+\sum_{j=1}^{N_p} \textcolor{black}{e_{t}^{dp_{j}}}+\sum_{j=1}^{N_p}e_{t}^{ \textcolor{black}  {movp_{j}}}\Big). \label{total_cost_moving2} 
\end{align}

 Here, \textcolor{black}{$e^{dq}_{t}$} represents the cost associated with the dynamic factor of the evader, while \textcolor{black}{$e_t^{dp_{j}}$} denotes the dynamic factor cost of the $j^{th}$ pursuer. The measurement factor costs are denoted by \textcolor{black}{$e^{mp_{j}}_{t}$} and \textcolor{black}{$e_t^{op_{ij}}$}. Specifically, $e^{mp_{j}}_{t}$ indicates the measurement factor cost between the $j^{th}$ pursuer and the evader, whereas \textcolor{black}{$e_t^{mo_{ji}}$} represents the measurement factor cost between the $i^{th}$ obstacle and the $j^{th}$ pursuer. \textcolor{black}{$e_{t}^{movp_{j}}$} represents the movement cost for the $j^{th}$ pursuer, minimizing the distance between the evader's estimated position and the $j^{th}$ pursuer. The terms \textcolor{black}{$e_t^{op_{ji}}$} and \textcolor{black}{$e_t^{cp_{ji}}$} correspond to the cost of the obstacle avoidance and the collision avoidance factor, respectively.
\begingroup\color{black}{
The overall cost function is iterating through all factors and the aim is to minimize the overall cost function over all factors in the formulated factor graph at time $T$.}\endgroup\ The problem consists of seven types of factors, described below:
\subsubsection{Dynamic Factor of Evader}
The evader aims to reach a predefined goal ($G$) by applying the DWA approach. The dynamic cost of the evader's movements at time step $t$ can be computed using the following cost function:
\begin{align}     \textcolor{black}{e_{t}^{dq}}(x_{t}^{q}) &= {||x_{t}^{q}-x_{t-1}^{q}||}_{\Omega_t^q}^2~ \label{dynamic_factor_evader}, \end{align} 
\textcolor{black}{where, $||.||$ denotes the Mahalanobis distance \cite{mclachlan1999mahalanobis}, and $\Omega_t^q$ represents the information matrices of the variables.} In Eq. \eqref{dynamic_factor_evader}, $x^q_{t}$ represents the evader's position at time $t$. The cost function measures the prediction error of the relative transformation between two consecutive evader poses, comparing the predicted motion based on current pose estimates with the actual odometry measurements. This error is weighted by the uncertainty of those measurements. 
\subsubsection{Dynamic Factor of Pursuers}
Similar to the cost function of the dynamic factor for the evader, the cost function for the dynamic factor of each pursuer can be defined as follows: 
\begin{align} \textcolor{black}{e_{t}^{dp_{j}}}(x_{t}^{p}) &= {||x_{t}^{p}-x_{t-1}^{p}||}_{\Omega_t^p}^2~\label{dynamic_factor_pursuer}, \end{align} 
where $x_{t}^{p}$ represents the pursuer's pose at time $t$.
\subsubsection{Measurement factor between each pursuer and evader}
The measurement factor between each pursuer and evader compares the estimated pose of the evader with the measured value. The measurement factor can be defined as follows: \begin{align} e_t^{mp_{j}}(x^{q}_{t},x^{p}_{t}) &= {||\hat{x}^q_{t}-h({x}^q_{t},{x}^p_{t})||}_{\Omega_t^{pq}}^2~\label{measurement_Factor}, \end{align} where $h$ represents the measurement function, and $\hat{x}^{q}_{t}$ is the evader's estimated position at time $t$.

\subsubsection{Measurement Factor between Pursuers and Obstacles}
    Pursuers need to localize themselves in the scene. 
    \textcolor{black}{To do this, the assumption is that a map of the environment is available for the pursuers in advance, and the pursuers utilize the map, along with the real-time measurement of the obstacles, to localize themselves. The  static obstacles' locations are assumed to be known and are used as landmarks for the localization of the robots. However, the target of the evader, as well as its position and orientation, are assumed to be unknown to the pursuers. The localization is performed via the proposed factor graph.} \\
\begingroup\color{black}{It is assumed that the pursuers have full observability of the scene and know the poses of the obstacles. They need to localize their own poses with respect to the obstacles.}\endgroup{} This measurement factor can be defined as follows: \begin{align} \textcolor{black}{e_t^{mo_{ji}}}(x^o,x^{p}_{t}) &= {||\hat{x}^p_{t}-h({x}^o,{x}^p_{t})||}_{\Omega_t^{qo}}^2~\label{measurement_Factor_obs}, \end{align} where $\hat{x}^p_{t}$ represents the estimated position of a pursuer at time $t$,  $x^{o}$ represents the position of obstacle $o$, \textcolor{black}{and $h(\cdot)$ is the measurement model which measures the bearing and range between a pursuer and obstacles~\cite{factor-graphs-for-perception}.}
\subsubsection{Planning Factor for Pursuers}
As seen in Fig.~\ref{pursuers_moving}, pursuers can move around the search space. The goal function defined for the pursuers is to get as close to the evader as possible. As mentioned earlier, the goal of the pursuer is to get as close to the evader as possible. To achieve this goal, each pursuer attempts to move as close as possible to $\hat{x}^{q}_{t}$, noted with
\begin{align}
    \textcolor{black}{e_{t}^{movp_{j}}}(\hat{x}^{q}_{t},\hat{x}_{t}^{p}) &= {||\hat{x}^{q}_{t}-\hat{x}_{t}^{p}||}_{\Omega_t^{pq}}^2~ \label{plan_factor}.
\end{align}

\subsubsection{Collision Avoidance Factor for Pursuers}
Since there is a possibility of collisions between pursuers, the collision avoidance factor is defined as follows:
\begin{align}
    \textcolor{black}{e_t^{cp_{ji}}}(x^{p}_{t}) &= \left\{\begin{aligned}
        {||1-\frac{d_{p}}{c_{1}}||}_{\Omega_t^{p}}^2~,&& \quad  d_p< d_{s} \\
        0,&&  d_p  \geq d_{s} 
    \end{aligned}
    \right. \label{collision_Factor}, 
\end{align}
where the objective is to define a safety bubble for each pursuer. The radius of the safety bubble is given by $d_s$. If the distance between pursuers $p_i$ and $p_j$ is less than $d_s$, the error term $||1-\frac{d_{p}}{c_{1}}||$ is applied. Here, \begingroup\color{black}{$c_1 = 2r_R + \epsilon$, where $\epsilon$ is a small safety distance and $r_R$ represents the robot's external body radius (see Table~\ref{vars_avoidance_extended} for values)}\endgroup{}, and $d_{p}$ represents the distance between the two pursuers, $p_i$ and $p_j$. 
\subsubsection{Obstacle Avoidance Factor for Pursuers}
To avoid collisions with obstacles in the scene, pursuers use the obstacle avoidance factor. Similar to the collision avoidance factor, the purpose of this factor is to create a safety bubble around the pursuers, preventing collisions with obstacles.
\begin{align}
    e_t^{op_{ji}}(x^{p}_{t}) &= \left\{\begin{aligned}
        {||1-\frac{d_{o}}{c_{2}}||}_{\Omega_t^{o}}^2~,&& \quad  d_o< d_{s} \\
        0,&&  d_o  \geq d_{s}  
    \end{aligned}    \right. 
    \label{obs_factor}.
\end{align}
Here, $d_{o}$ represents the distance between each moving pursuer and the obstacles in the scene, while \begingroup\color{black}{$c_2 = r_R$  (see Table~\ref{vars_avoidance_extended} for values)}\endgroup{}.
The optimized cost function generates the path based on the estimated position of pursuers ($\hat{x}^p_{t}$) and the evader ($\hat{x}^q_{t}$) at time $t$. 

\subsection{Time Complexity Analysis}
To analyze the complexity of the problem, the worst-case scenario is considered. The number of variables and factors grows over time. The number of variables added at each time step is $N_p+N_q$.
To investigate the number of factors added at each time step, the effect of each factor must be investigated. The number of factors added at each time step for the dynamic factor of pursuers $f^{dp}$, planning factor of pursuers \textcolor{black}{$f^{movp}$}, and measurement factor between pursuers and evader $f^{mp}$ is the same and is $N_p$. The number of factors added for the dynamic factor of evader ($f^{dq}$) at each time step is $N_q$. The number of factors added at each time step for collision avoidance factor is  $\frac{N_{p} \times (N_{p} - 1)}{2}$. The number of factors added for the measurement factor between pursuers and obstacles ($f^{mo}$) and obstacle avoidance factor$(f^{op})$ is $N_p \times N_o$.
\\
The overall complexity is 
{
\begin{multline*}
O((N_{p} + N_{q}) \times n + (\frac{N_{p} \times (N_{p} - 1)}{2} 
+ 2 \times N_{p} \times N_{o}\\ 
+ N_{p} \times N_{q} + N_{p} + 1) \times n ),
\end{multline*}
}
The first part, which represents the number of variables, grows linearly in $O(n)$, while the second part, which represents the number of factors added at each time step, grows quadratically in $O(n^2)$. Thus, the overall complexity of the problem is in $O(n^2)$.

\begin{figure}[H]
    \centerline{\includegraphics[scale=0.35]{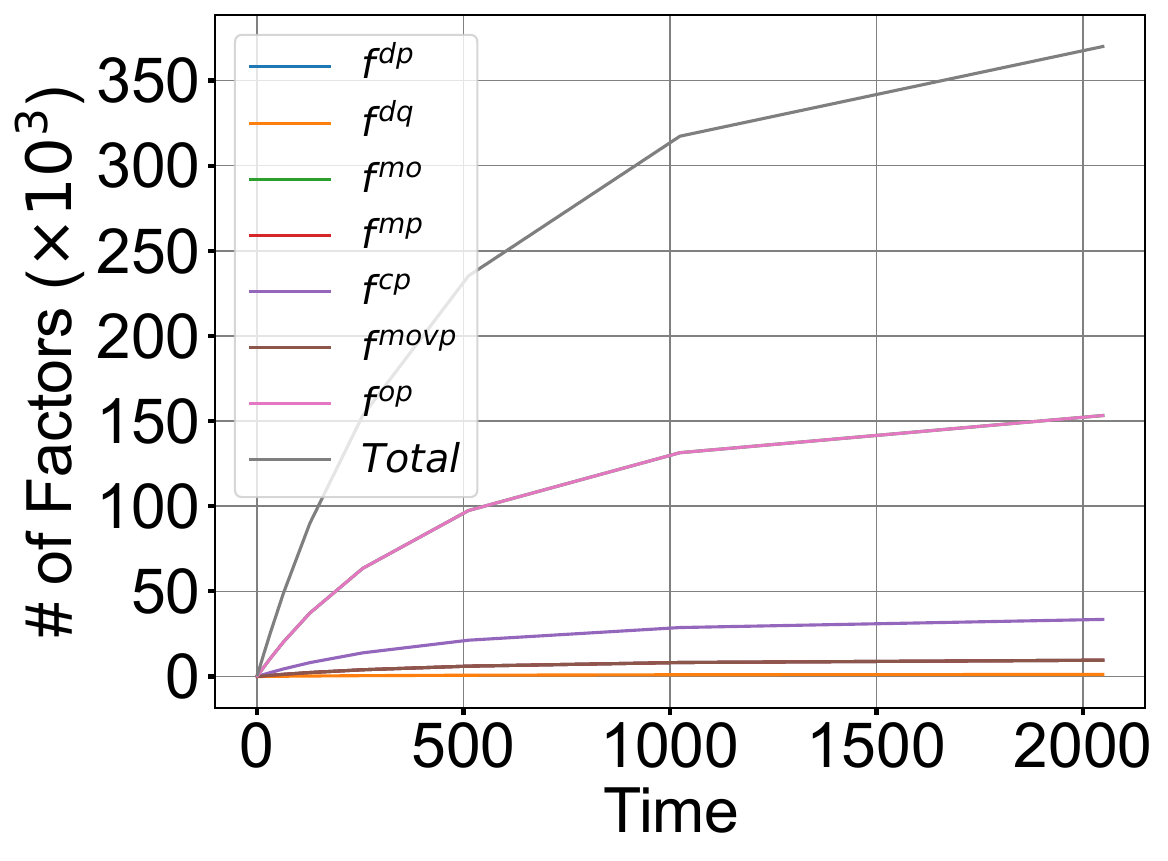}}
\caption{The number of different factors at each time step is shown in the figure above. In this figure, \(f^{op}\) and \(f^{mo}\) are increasing at the same rate, while \(f^{dp}\), \(f^{mp}\), and \(\textcolor{black}{f^{movp}}\) are increasing at the same speed.
}
    \label{number_factors}
\end{figure}
As shown in Fig.~\ref{number_factors}, the number of measurement factors between pursuers and obstacles and obstacle avoidance factors increase faster compared with other factors and variables.
The LM solver can be approximated by $\widetilde{O}(\epsilon^{-2})$~\cite{bergou2020convergence} where $\widetilde{O}$ exhibits the existence of logarithmic factors in $\epsilon$, \textcolor{black}{and $\epsilon$ represents an $\epsilon$-stationary point}. By considering $m$ as solver parameters and considering the post-optimization that takes place in $O(I)$, the overall complexity would be $\widetilde{O}(n \epsilon^{-2} (n^2 + m) + nI)$.
\subsection{Convergence Analysis}
To analyze the convergence of the proposed method, the inference method must be investigated, and a scenario that always converges should be provided.  Referring to Eq.~\eqref{total_cost_moving2}, the cost function consists of nonlinear least square parts. It is proven that MAP inference becomes the maximization of the product of all factors~\cite{dellaert2021factor} as 
\begin{align}
X^{\text{MAP}} = \arg\max_X \prod_i f_i(X_i)
\label{MAP_min},
\end{align}
where factors are in the following form 
\begin{align}
f_i(X_i) \propto \exp \left\{ -\frac{1}{2} \left\| h_i(X_i) - z_i \right\|_{\textcolor{black}{\Omega}}^2 \right\}.\label{mean_MAP_min}
\end{align}

In Eq.~\eqref{mean_MAP_min}, $\Sigma_{i}$ shows the covariance, $z_{i}$ shows the actual measurement, and $h_{i}(X_{i})$ shows the measurement model. 
Taking a negative likelihood of Eq.~\eqref{mean_MAP_min}, the problem becomes the minimization of non-linear factors:
\begin{align}
X^{\text{MAP}} = \arg\min_X \sum_i \left\| h_i(X_i) - z_i \right\|_{\Omega_i}^2.
\label{drop_mean_MAP_min}
\end{align}

The LM optimizer can solve Eq.~\eqref{drop_mean_MAP_min} by linearizing it as:
\textcolor{black}{
\begin{align}
\Delta^{*}=\arg\min_{\Delta} (A\Delta - b)^\top (A\Delta - b).
\label{linear_drop_mean_MAP_min}
\end{align}}

In Eq.~\eqref{linear_drop_mean_MAP_min}, $A$ is the single measurement Jacobian, and $b$ shows all the prediction errors in the form of $z_{i}-h_{i}(X^{0}_{t})$. \textcolor{black}{ $\Delta$ denotes the state increment, the change in the evader's pose at time $t$.}

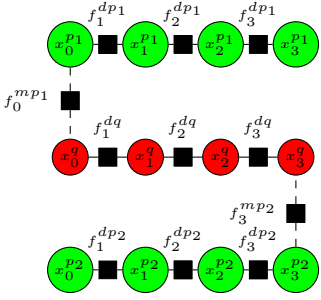
\begin{figure}[t!]
\centering
\begin{tikzpicture}[scale=1]
    \foreach \pos/\name/\color in {{(0,0)/x_{0}^q/red}, {(2,0)/x_{1}^q/red},{(4,0)/x_{2}^q/red},{(6,0)/x_{3}^q/red},{(0,3)/x_{0}^{p_{1}}/green},{(2,3)/x_{1}^{p_{1}}/green},{(4,3)/x_{2}^{p_{1}}/green},{(6,3)/x_{3}^{p_{1}}/green},{(0,-3)/x_{0}^{p_{2}}/green},{(2,-3)/x_{1}^{p_{2}}/green},{(4,-3)/x_{2}^{p_{2}}/green},{(6,-3)/x_{3}^{p_{2}}/green}}
\node[circle,draw,fill=\color,minimum size=.25cm,inner sep=1pt,font=\tiny] (\name) at \pos {$\name$}; 
    \foreach \pos/\name in {{(1,0)/1},{(3,0)/3},{(5,0)/5},{(0,1.5)/f3},{(6,-1.5)/f10},{(1,3)/f11},{(1,3)/f12},{(3,3)/f13},{(5,3)/f14}}
        \node[rectangle,draw,fill=black,minimum size=0.1cm] (\name) at \pos {}; \foreach \pos/\name in {{(3,-3)/f16},{(5,-3)/f17}}
        \node[rectangle,draw,fill=black,minimum size=0.05cm] (\name) at \pos {}; 
    \foreach \pos/\name/\label/\text in {{(1,-3)/f15/$f_1^{dp_{2}}$},{(3,-3)/f16/$f_2^{dp_{2}}$},{(5,-3)/f17/$f_3^{dp_{2}}$},{(1,3)/f12/$f_1^{dp_{1}}$},{(3,3)/f13/$f_2^{dp_{1}}$},{(5,3)/f14/$f_3^{dp_{1}}$},{(1,0)/1/$f_1^{dq}$},{(3,0)/3/$f_2^{dq}$},{(5,0)/5/$f_3^{dq}$}}
        \node[rectangle,draw,fill=black,minimum size=0.1cm,label={[font=\tiny]90:\text}] (\name) at \pos {}; 
    \foreach \pos/\name/\label/\text in {{(0,1.5)/$f_{0}^{mp_{1}}$},{(6,-1.5)/$f_{3}^{mp_{2}}$}}
        \node[rectangle,draw,fill=black,minimum size=0.1cm,label={[font=\tiny]180:\text}] (\name) at \pos {}; 
    \foreach \source/\target/\color in {x_{0}^{p_{1}}/f11/black,f11/x_{1}^{p_{1}}/black,x_{1}^{p_{1}}/f13/black,f13/x_{2}^{p_{1}}/black,x_{2}^{p_{1}}/f14/black,f14/x_{3}^{p_{1}}/black}
        \path (\source) edge[draw=\color] (\target);
    \foreach \source/\target/\color in {x_{0}^{p_{2}}/f15/black,f15/x_{1}^{p_{2}}/black,x_{1}^{p_{2}}/f16/black,f16/x_{2}^{p_{2}}/black,x_{2}^{p_{2}}/f17/black,f17/x_{3}^{p_{2}}/black}
        \path (\source) edge[draw=\color] (\target);
    \foreach \source/\target/\color in {x_{0}^q/1/black, x_{1}^q/1/black, x_{1}^q/x_{2}^q/black, x_{2}^q/x_{3}^q/black}
        \path (\source) edge[draw=\color] (\target);
            \foreach \source/\target/\color in {x_{0}^{p_{1}}/f3/black,f3/x_{0}^q/black,
            x_{3}^{p_{2}}/f10/black,f10/x_{3}^q/black}
        \path (\source) edge[draw=\color,dashed] (\target);       
\end{tikzpicture}
\caption{A simplified factor graph with a tree structure is presented. In this scenario, there is only one measurement from each pursuer and evader. 
}
\label{pursuers_moving2}
\end{figure}

As a simple scenario, if there are two pursuers, $p_{1}$ and $p_{2}$, and they observe the evader only once the factor graph simplifies as shown in Fig.~\ref{pursuers_moving2}. This graph has a tree structure, and it converges surely~\cite{koller2009probabilistic}. 
In the factor graph, the factor function for the movements of pursuers is denoted as $f_1^{dp_{1}}$ for $p_1$ at time-step 0, as shown in Fig.~\ref{pursuers_moving2}. The prior factor of the pursuer $p$ is represented as $f_{0}^{p_1}$ at time-step 0. Variable elimination is an exact inference method.
As an example, \textcolor{black}{the probability function of variable $x_3^q$} can be found by summing out all other variables in the joint distribution:
\textcolor{black}{
\begin{align}
P(x^q_{3}) = & \sum_{x_{0}^{q}} \cdots \sum_{x_{3}^{q}} \sum_{x_{0}^{p_{1}}} \cdots \sum_{x_{3}^{p_{1}}} \nonumber \\
& \sum_{x_{0}^{p_{2}}} \cdots \sum_{x_{3}^{p_{2}}} P(x^q_{0}, \ldots, x^q_{3}, x_{0}^{p_{1}}, \ldots, x_{3}^{p_{1}}, x_{0}^{p_{2}}, \ldots, x_{3}^{p_{2}}).
\label{probability}
\end{align}}
\textcolor{black}{In Eq. \eqref{probability}, all variables are marginalized out, except $x_3^{q}$.} By starting the elimination process in the graph shown in Fig.~\ref{pursuers_moving2}, we begin by eliminating node $x_{0}^{p_{1}}$. The involved factors are $f_{0}^{p_{1}}$, ${f_{0}^{mp_{1}}}$, and ${f_{1}^{dp_{1}}}$. After eliminating $x_{0}^{p_{1}}$, two marginal factors are created:
\begin{align}
F(x_{0}^{p_{1}},f_{1}^{dp_{1}})=f_{0}^{p_{1}}. f_{1}^{dp_{1}}
\label{fi1}
\end{align}
\begin{align}
F(x_{0}^{p_{1}},f_{0}^{mp_{1}})=f_{0}^{p_{1}}.f_{0}^{mp_{1}}
\label{fi2}
\end{align}

By marginalizing over $x_{0}^{p_1}$, impacted factors are:
\begin{align}
f_{1}^{d_{p_{1}}} = \sum_{x_{0}^{p_{1}}} F(x_{0}^{p_{1}}, f_{1}^{d{p_{1}}}).
\label{marg_fi1}
\end{align}
\begin{align}
f_{0}^{mp_{1}}=\sum_{x_{0}^{p_{1}}}F(x_{0}^{p_{1}},f_{0}^{mp_{1}}).
\label{marg_fi2}
\end{align}

\textcolor{black}{In the next steps, we can remove the nodes representing the evader's position, starting by eliminating the node \(x^{q}_0\). The updated factor is then only \(f_0^{dq}\), which can be computed as follows:
\\
\begin{align}
F(x_{0}^{q},f_{0}^{dq})=f_{0}^q.f_{1}^{dq}
\label{fi3}
\end{align}
By marginalizing over $x_{0}^{q}$, impacted factors will be:
\begin{align}
f_{1}^{d_q} = \sum_{x_{0}^q} F(x_{0}^{q}, f_{1}^{dq})
\label{marg_fi3}
\end{align}
} 
The elimination step, continues for nodes $x_1^q$, $x_2^q$, $x_3^q$, and$x_3^{p_2}$ in the same manner.  
The starting point is not important since there is no cycle in the tree. \textcolor{black}{In conclusion, if the graph does not have a cycle, it surely converges. Conversely, if it has a cycle, MAP inference converges most of the time~\cite{factor-graphs-for-perception}.}

\subsection{Uncertainty Minimization Analysis}
\textcolor{black}{In this section, we investigate the effects of different factors on the uncertainty of predictions and the impact of measurements on that uncertainty.
If we consider the evader to be moving around the search space, the new pose of the evader can be assessed using the following equation:
\begin{equation}
    x_{t}^q = f(x^q_t, \Delta x^q_t) + \Sigma_t^q.
    \label{eq:motion_model}
\end{equation}
In Eq.~\eqref{eq:motion_model}, $f$ is the motion model function that moves the evader by $\Delta x^q_t$.
 Given a sequence of odometry readings, the LM optimizer aims to minimize the sum of the squared residuals between the predicted and actual robot poses as shown in Eq. \eqref{dynamic_factor_evader} \cite{marquardt1963algorithm}. The update step is performed using the following equation: 
\begin{equation}
x_{t}^{q} = x_{t-1}^{q} - (J^T J + \lambda I)^{-1} J^T e_{t}^{dq}.
\label{eq:lm_update}
\end{equation}
In the above equation, \(J\) represents the Jacobian matrix of the cost function, and \(\lambda\) is the damping factor that balances between the gradient descent and Gauss-Newton methods.
}
\textcolor{black}{By considering that the only factor existing in the problem is the dynamic factor, the uncertainty can be expressed as:
\begin{equation}
    \sigma_T = \sigma_0 + \sum_{i=0}^T e_t^{dq} - \Delta_{LM}
    \label{final_uncertainty}
\end{equation}
This equation represents the accumulated uncertainty. In Eq.~\eqref{final_uncertainty}, \(\sigma_0\) denotes the uncertainty at time 0, and \(\Delta_{LM}\) represents the correction from the LM optimizer.
}

To minimize the uncertainty, a measurement must be obtained to correct the estimations. When a measurement arrives from the pursuers, the pose of the evader can be estimated as follows:
\begin{equation}
x_{t}^{q} = x_{t-1}^{q} - (J^T J + \lambda I)^{-1} J^T e_t^{mp_{j}}.
\label{uncertainty_measurement}
\end{equation}

Now, the Jacobian is updated, incorporating richer information, which improves the conditioning of the problem. This leads to more accurate estimations and reduced uncertainty. Similarly, the dynamic factor of pursuers (\(f^{dp}\)) and the measurement factors between pursuers and landmarks (\(f^{mo}\)) can also be modeled, contributing to the reduction of uncertainty.
\section{Experiments}\label{sec:results}
Experiments are conducted in two environments: (i) a simulated environment and (ii) a real-world hardware environment. The performance analysis is based on the average distance traveled by pursuers, time, and the number of time steps. \textcolor{black}{In our study, to ensure fairness and consistency, we employed an exhaustive search over the parameters and weight space for the same scenario. This approach allows us to rigorously evaluate different weight configurations and verify that the reported results are not an artifact of arbitrary parameter selection, but rather reflect robust performance across a range of weight choices. In total, we evaluated  1000 combinations of input sets. In order to consider various scenarios, we vary the values related to dynamic factor uncertainties, measurement uncertainty, and the frequency of measurements.\newline
To evaluate the performance of different weight settings, we measured the time required for four pursuers to capture an evader moving from left to right.
\begin{figure}[H]
    \centerline{\includegraphics[scale=0.35]{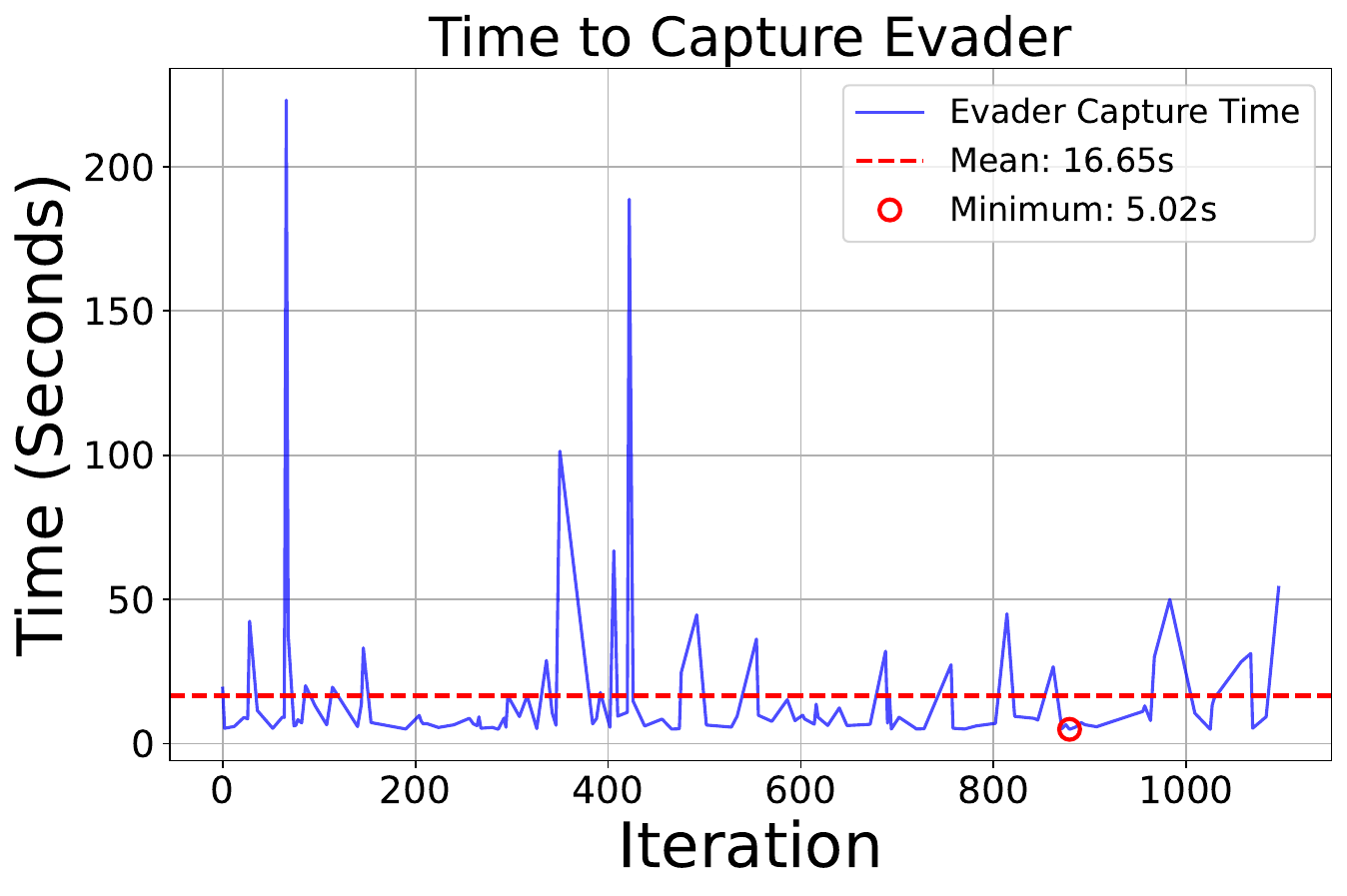}}
    \caption{\textcolor{black}{Time to capture the evader for four pursuers in  1000 iterations to find the optimal weights for our problem.} 
    }
    \label{exhaustive_search}
\end{figure}
As shown in Fig.~\ref{exhaustive_search}, the optimized weights catch the evader in 5 seconds, which is the minimum time to capture the evader over 1000 iterations.  
In order to show the correlation between each parameter and time to capture the evader the correlation analysis was conducted.
\begin{figure}[H]
    \centerline{\includegraphics[scale=0.35]{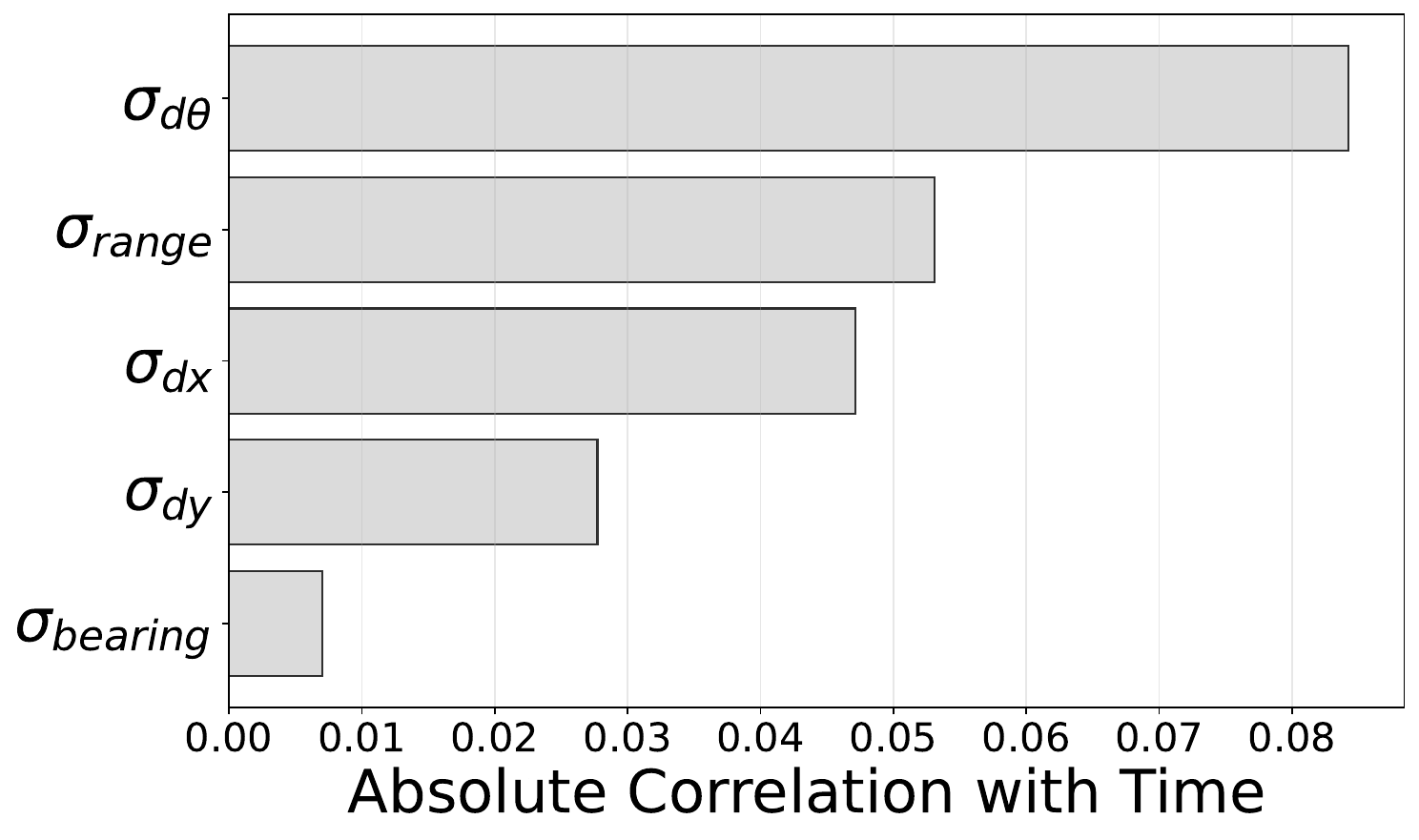}}
    \caption{\textcolor{black}{Correlation analysis between parameters and time to capture the evader (See Table~\ref{vars_avoidance_extended} for parameters).}}
    \label{time_bars}
\end{figure}
In Fig.~\ref{time_bars}, the correlation between the time to capture the evader and the parameters is shown. The correlation quantifies the strength and direction of the linear relationship between each parameter and the capture time. It is obtained by computing the Pearson correlation coefficient~\cite{benesty2009pearson} across multiple simulation runs, where each run varies one or more parameters while recording the corresponding time required to capture the evader. A positive correlation indicates that increasing the parameter tends to increase the capture time, while a negative correlation indicates that increasing the parameter tends to reduce the capture time.
The obtained optimal values utilizing exhaustive search are as follows:
}

\begin{itemize}
  \setlength{\parskip}{0pt}
  \setlength{\itemsep}{0pt plus 1pt}
    \item {Dynamic factor:}
    In this problem, the uncertainties for each step along the x-axis and y-axis are denoted as $\sigma_{\textcolor{black}{dx}}$ = 0.1 and $\sigma_{\textcolor{black}{dy}}$ = 0.1, respectively. Additionally, the uncertainty in orientation is $\sigma_{\textcolor{black}{d\theta}}$ = 0.01. 
    \item {Measurement factor:}
The uncertainty of each measurement factor is represented by $\sigma_{\text{range}}$ and $\sigma_{\text{bearing}}$. In this context, $\sigma_{\text{range}}$ typically refers to the uncertainty associated with the distance measurement, while $\sigma_{\text{bearing}}$ represents the uncertainty in the angle or direction measurement.
For the simulation, we set $\sigma_{\text{range}} = 10$ and $\sigma_{\text{bearing}} = 0.05$ for both $f_{t}^{mo}$ and $f_{t}^{mp}$. 
\item {Prior factor:}
Since agents are in motion, it is necessary to designate certain factors as prior factors to establish reference points and enhance measurement accuracy. In this context, 'prior factors' refer to fixed reference points or landmarks.
\item {Collision avoidance factor:}
In order to prevent pursuers from having collisions, this factor is utilized. The uncertainties for each step along the x-axis and y-axis are denoted as $\sigma_{\textcolor{black}{cpx}}$ = 1 and $\sigma_{\textcolor{black}{cpy}}$ = 0.1, respectively.
\item {Obstacle avoidance factor:}
In order to prevent pursuers having collision with obstacles, this factor is utilized. The uncertainties for each step along the x-axis and y-axis are denoted as $\sigma_{\textcolor{black}{opx}}$ = 1 and $\sigma_{\textcolor{black}{opy}}$= 0.1 respectively.
\end{itemize}
\textcolor{black}{Table~\ref{vars_avoidance_extended} lists the tuned parameters used for the simulated experiments.} 
\begin{table}[h!]
\centering
\scriptsize
\color{black} 
\begin{tabular}{l l l}
\toprule
\textbf{Variable} & \textbf{Definition} & \textbf{Value} \\
\midrule
$r_R$                & Pursuer external body radius & 0.3 m \\
$c_1$                & $c_1 = 2 r_R + \epsilon$ & 0.61 m \\
$\epsilon$                & Safety Margin & 0.1 m \\
$c_2$                & $c_2 = r_R$ & 0.3 m \\
$d_s$                & Safety distance bubble & 0.6 m \\
$\sigma_{dx}$  & Uncertainty along x-axis in dynamic factor for pursuers and evader & 0.1 m\\
$\sigma_{dy}$ & Uncertainty along y-axis in dynamic factor for pursuers and evader & 0.1 m\\
$\sigma_{d\theta}$      & Orientation uncertainty in dynamic factor for pursuers and evader & 0.01 rad \\
$\sigma_\text{range}$ & Range measurement uncertainty in measurement factor & 10 m\\
$\sigma_\text{bearing}$ & Bearing measurement uncertainty in measurement factor & 0.05 rad\\
$\sigma_{cpx}$ & Uncertainty along x-axis in collision avoidance factor & 1 m\\
$\sigma_{cpy}$  & Uncertainty along y-axis in collision avoidance factor & 0.1 m \\
$\sigma_{opx}$ & Uncertainty along x-axis in obstacle avoidance factor & 1 m\\
$\sigma_{opy}$   & Uncertainty along y-axis in obstacle avoidance factor & 0.1 m \\
\bottomrule
\end{tabular}
\vspace{-3 mm}
\caption{\textcolor{black}{List of parameters used for the experiment.}}
\label{vars_avoidance_extended}
\end{table}
\subsection{Simulation}
The simulated environment is developed in a way that the number of pursuers, evaders, and obstacles in the scene can be modified. 
The simulated environment is developed in a  $3500 \times 3500$ window size. Additionally, the simulated environment is capable of simulating different trajectories of the evader. The simulation is implemented in C++  and OpenCV library is utilized for visualization purpose.
\subsubsection{Comparison}
Various approaches have been explored in pursuit of strategies for capturing and chasing evaders. Among these, Qin \textit{et al.}~\cite{yang2022game} introduced a hierarchical network-based model aimed at devising a cooperative pursuit strategy. \textcolor{black}{It aims to solve the pursuit-evasion problem by generating utility trees (GUT)~\cite{yang2022game}.}  
    \textcolor{black}{To ensure a fair comparison, the initial poses and speeds of both the evader and pursuers are kept the same for the proposed and compared methods.} \textcolor{black}{Two additional tactics for pursuers}  
    are \textcolor{black}{also} considered: Constant Bearing (CB) and Pure Pursuit (PP) strategies~\cite{makkapati2019optimal}. In the CB strategy, the bearing angle between a pursuer and the evader remains fixed until the time of capture. Conversely, in the PP strategy, the pursuer's velocity vector aligns with the line of sight. Our proposed method operates at a frequency of 20 Hz on average, while the compared methods operate at a frequency of 30 Hz on average.
    \begin{figure}[H]
    \centering
    \begin{subfigure}[b]{0.49\linewidth}
        \includegraphics[width=\linewidth]{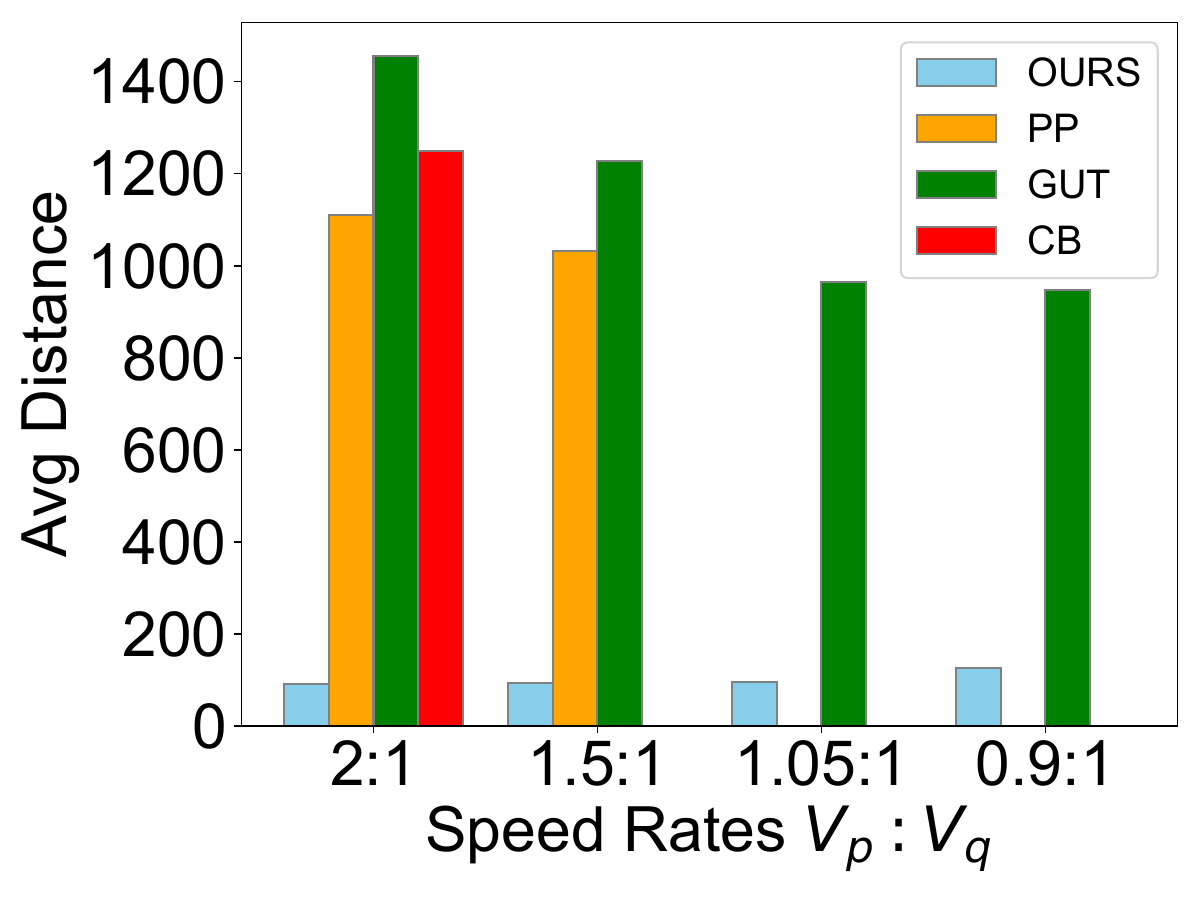}
        \label{fig:sub1}
        \vspace{-.5 cm}
    \end{subfigure}
    \begin{subfigure}[b]{0.49\linewidth}
        \includegraphics[width=\linewidth]{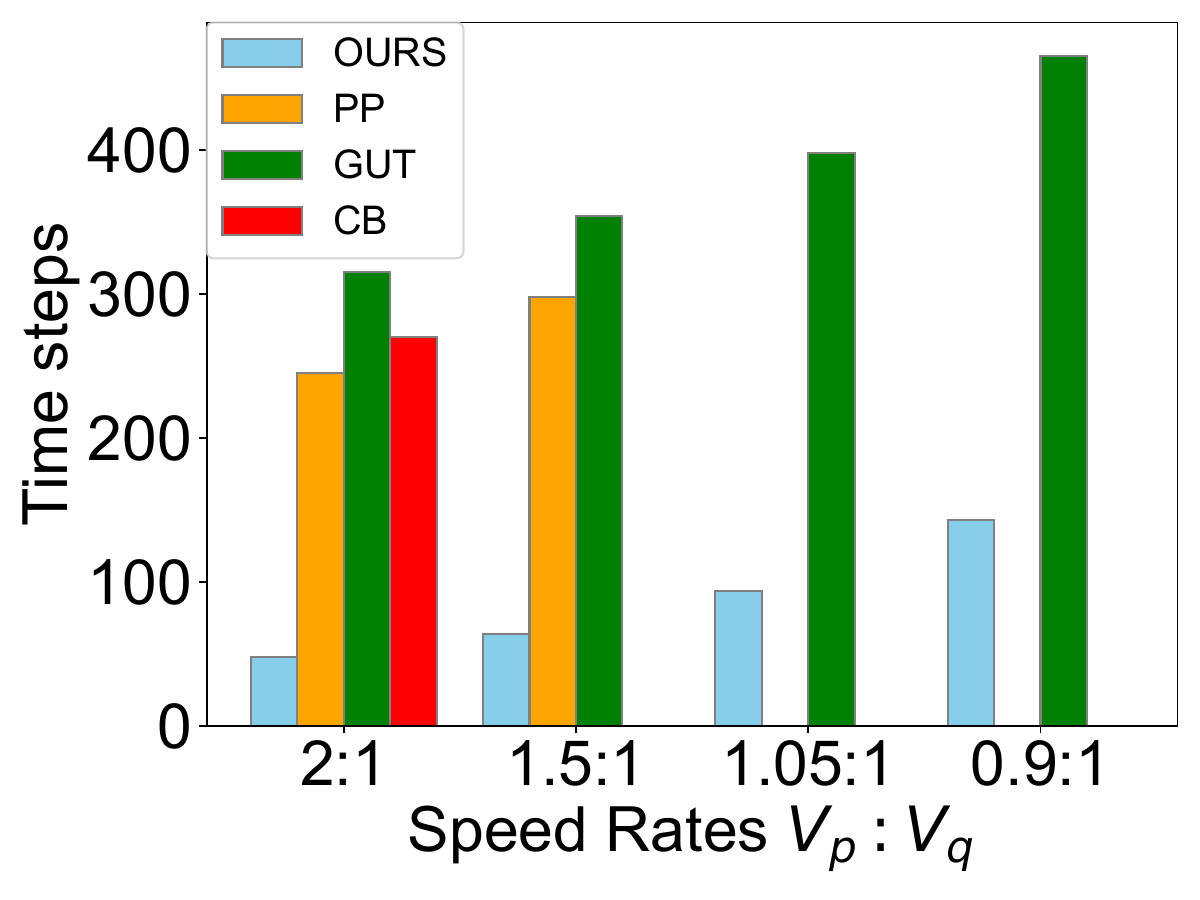}
        \label{fig:sub2}
        \vspace{-.5 cm}
    \end{subfigure}
    \vspace{0 mm}
    \caption{    Average distance traveled by pursuers (left) and time to capture the evader (right) for different approaches. $V_p$ and $V_{q}$ show the rate of pursuer speed in comparison to evader speed. PP and CB methods are not able to catch the evader at all speed rates, \textcolor{black}{in which no values are reported for them.} 
    } \label{fig:main_comparison2}
\end{figure}
As can be seen in Fig.~\ref{fig:main_comparison2}, the proposed method captures 
the evader faster in all speed rates that were tested, and also the average distance traveled by pursuers is much less in our method. In speed rates ($V_p$=1.05:$V_q$=1) and ($V_p=0.9:V_q=1$) pursuers fail to catch the evader in PP and CB methods, while our method can pass. PP was not able to catch the evader in half of the experiments and CB was not able to catch the evader in 75\% of experiments.  A key factor in our method is the rate of measurement coming from pursuers. In the experiments, it is set to 0.2 Hz for our proposed method, while in other methods it is 1 Hz. This means that the proposed method does not need to have measurements in all time steps and can lead to less memory consumption and computation overhead.
\subsubsection{Different Trajectories for Evader}
\begin{figure}[H]
  \centering

  \begin{minipage}[b]{0.40\linewidth}
    \centering
    \includegraphics[width=\linewidth]{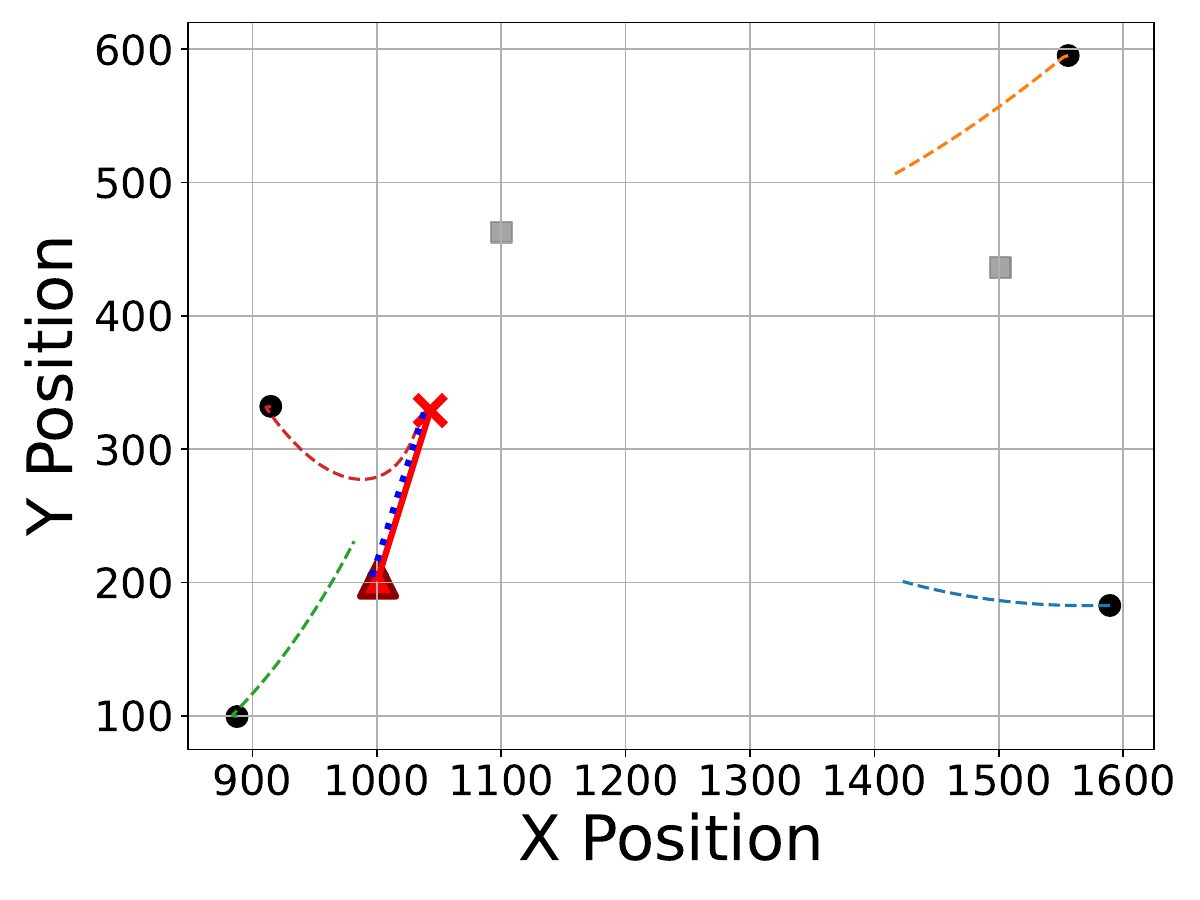}
  \end{minipage}\hfill
  \begin{minipage}[b]{0.40\linewidth}
    \centering
    \includegraphics[width=\linewidth]{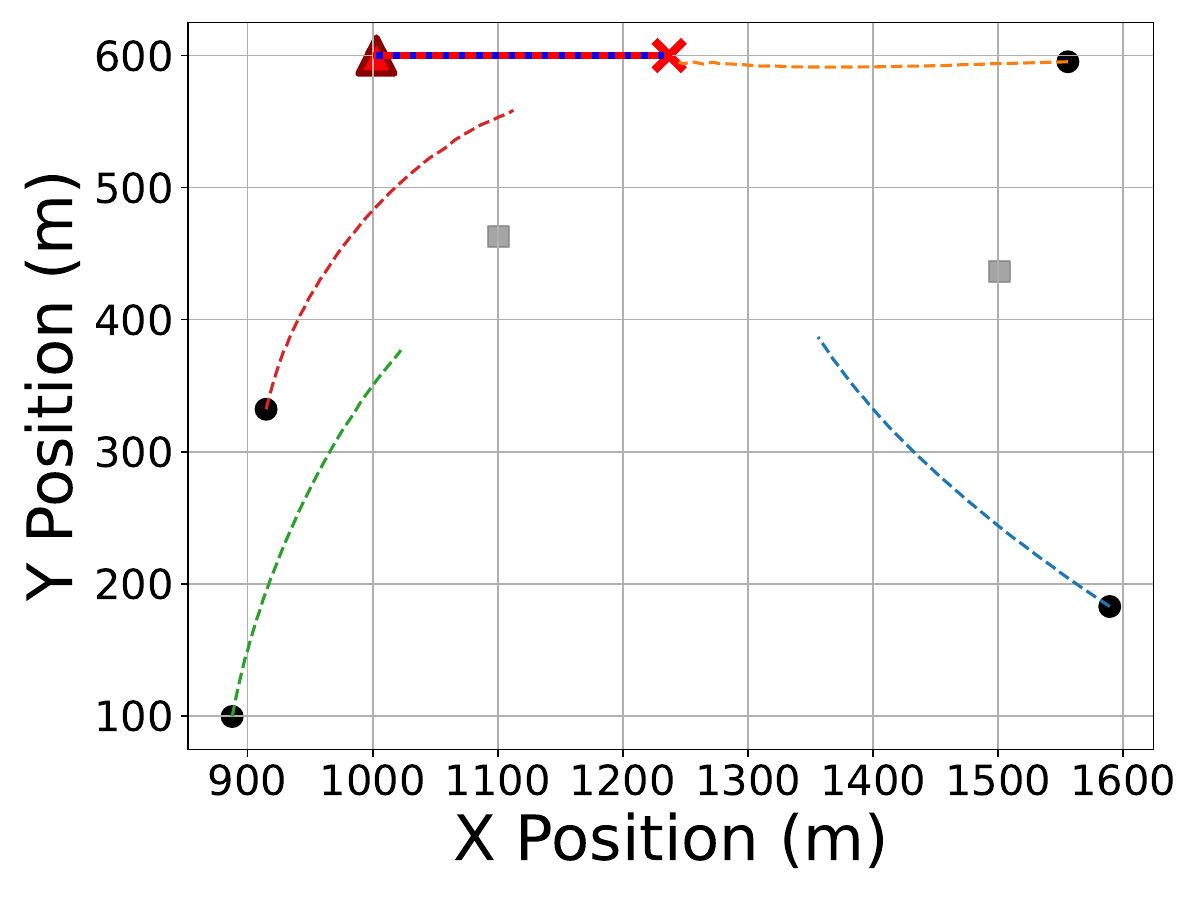}
  \end{minipage}

  \vspace{2mm}

  \begin{minipage}[b]{0.40\linewidth}
    \centering
    \includegraphics[width=\linewidth]{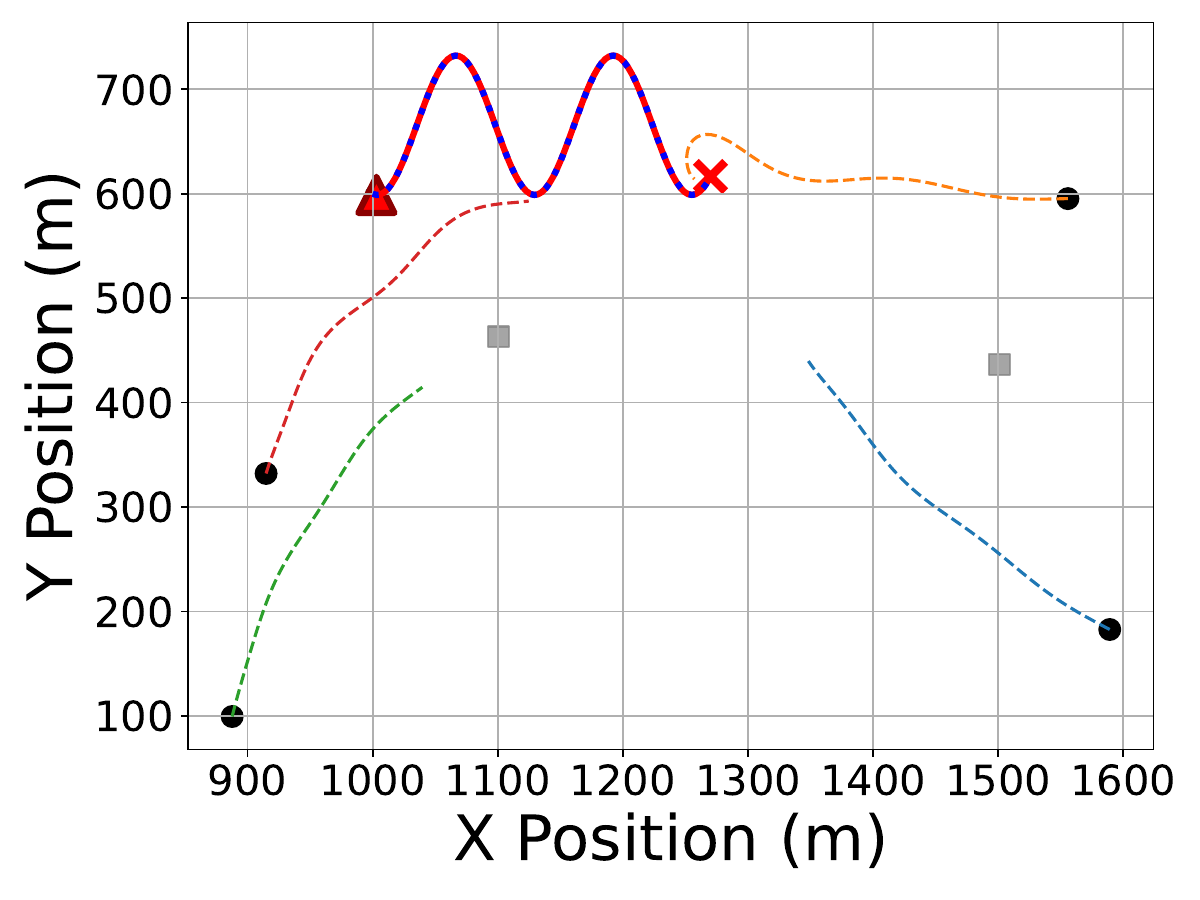}
  \end{minipage}\hfill
  \begin{minipage}[b]{0.40\linewidth}
    \centering
    \includegraphics[width=\linewidth]{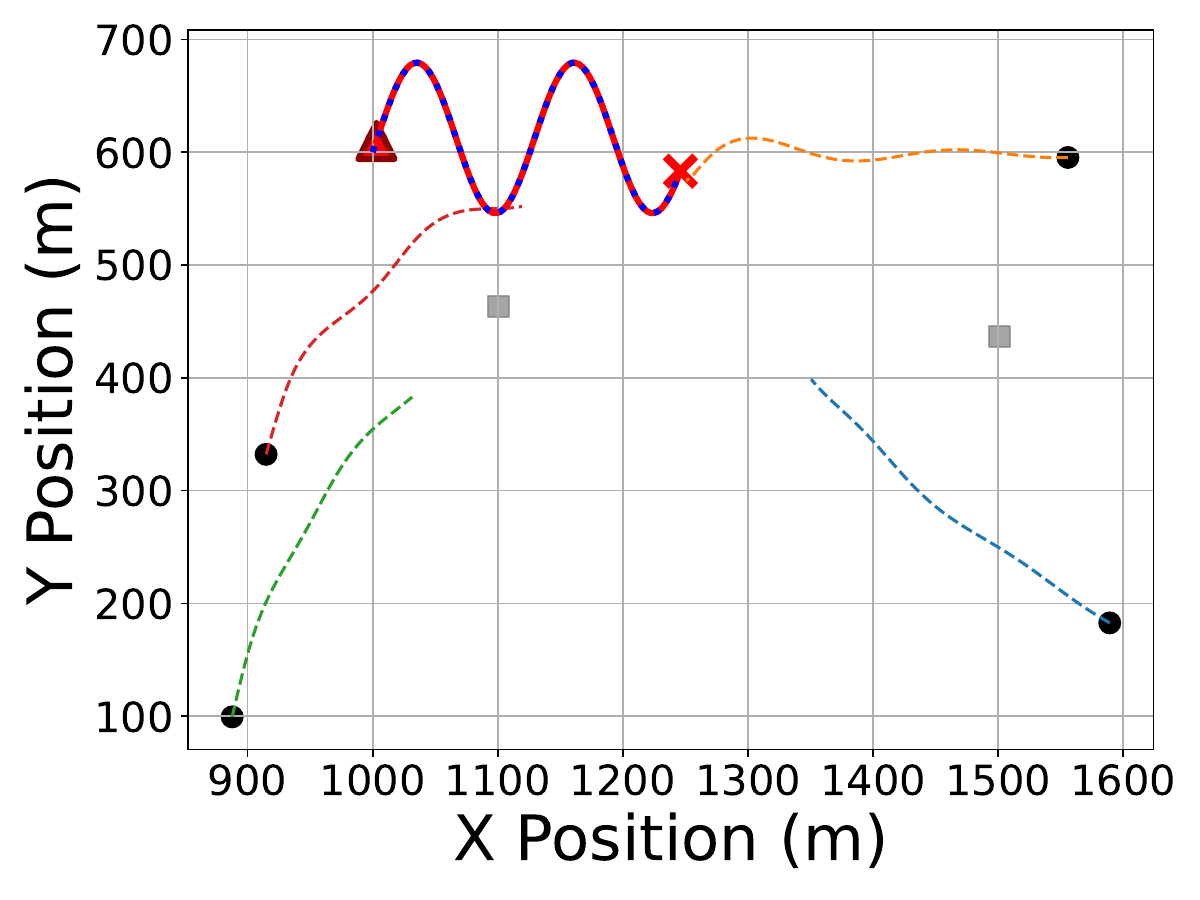}
  \end{minipage}

  \vspace{4mm}

  \setlength{\tabcolsep}{16pt}
  \begin{minipage}{0.95\linewidth}
    \centering
\newcommand{\legenditem}[2]{%
  \makebox[1.4cm][l]{#1} #2
}
\begin{tabular}{ll}
  \legenditem{\tikz[baseline=-0.6ex]\draw[line width=1pt, black, dashed] (0,0) -- (1.0cm,0);}{Pursuer 1} & 
  \legenditem{\tikz[baseline=-0.6ex]\draw[line width=1pt, red] (0,0) -- (1.0cm,0);}{Evader ground truth} \\

  \legenditem{\tikz[baseline=-0.6ex]\draw[line width=1pt, orange, dashed] (0,0) -- (1.0cm,0);}{Pursuer 2} & 
  \legenditem{\tikz[baseline=-0.6ex]\draw[line width=1pt, black, dotted] (0,0) -- (1.0cm,0);}{Evader estimate} \\

  \legenditem{\tikz[baseline=-0.6ex]\draw[line width=1pt, green, dashed] (0,0) -- (1.0cm,0);}{Pursuer 3} & 
  \legenditem{\tikz[baseline=-0.6ex]\filldraw[red] (0,0.06) -- (-0.05,-0.04) -- (0.05,-0.04) -- cycle;}{Evader start point} \\

  \legenditem{\tikz[baseline=-0.6ex]\draw[line width=1pt, red, dashed] (0,0) -- (1.0cm,0);}{Pursuer 4} & 
  \legenditem{\tikz[baseline=-0.6ex]\draw[red, thick] (-0.05,-0.05) -- (0.05,0.05) (-0.05,0.05) -- (0.05,-0.05);}{Evader final point} \\

  \legenditem{\tikz[baseline=-0.6ex]\filldraw[black] (0,0) circle (0.07);}{Pursuer start point} & 
  \legenditem{\tikz[baseline=-0.6ex]\filldraw[gray] (0,0) rectangle (0.12,0.12);}{Obstacles} \\
\end{tabular}
  \end{minipage}

   \caption{ The effect of different trajectories on the estimation over \begingroup\color{black}{1600}\endgroup{}  time-steps has been investigated. The black circle shows the start position of the pursuers. They try to get closer to the evader and catch it. \begingroup\color{black}{Dashed lines show their trajectory. The dotted black line represents the predicted trajectory of the evader, and the red line represents the evader’s actual (ground truth) trajectory. Obstacles are shown as gray squares, and the cross marks the evader’s final position. Utilizing FG-PE, the evader is captured when operating at a measurement frequency of 1 Hz. }\endgroup}
  \label{fig:trajectories}
\end{figure}

To investigate the effect of different trajectories on the performance of the proposed method, various evader trajectories were tested. As shown in Fig.~\ref{fig:trajectories}, the pursuers demonstrate accurate position estimation and effective pursuit of the evader. In the figure, the initial positions of the pursuers are indicated by circles, and there are four pursuers in the scene. \textcolor{black}{The measurement frequency is 1 Hz. The measurement frequency refers to the number of timesteps at which measurements are taken; a lower frequency means that measurements are obtained less often, which can increase uncertainty in the estimation process.}

\begin{figure}[h!]
    \centerline{\includegraphics[scale=0.35]{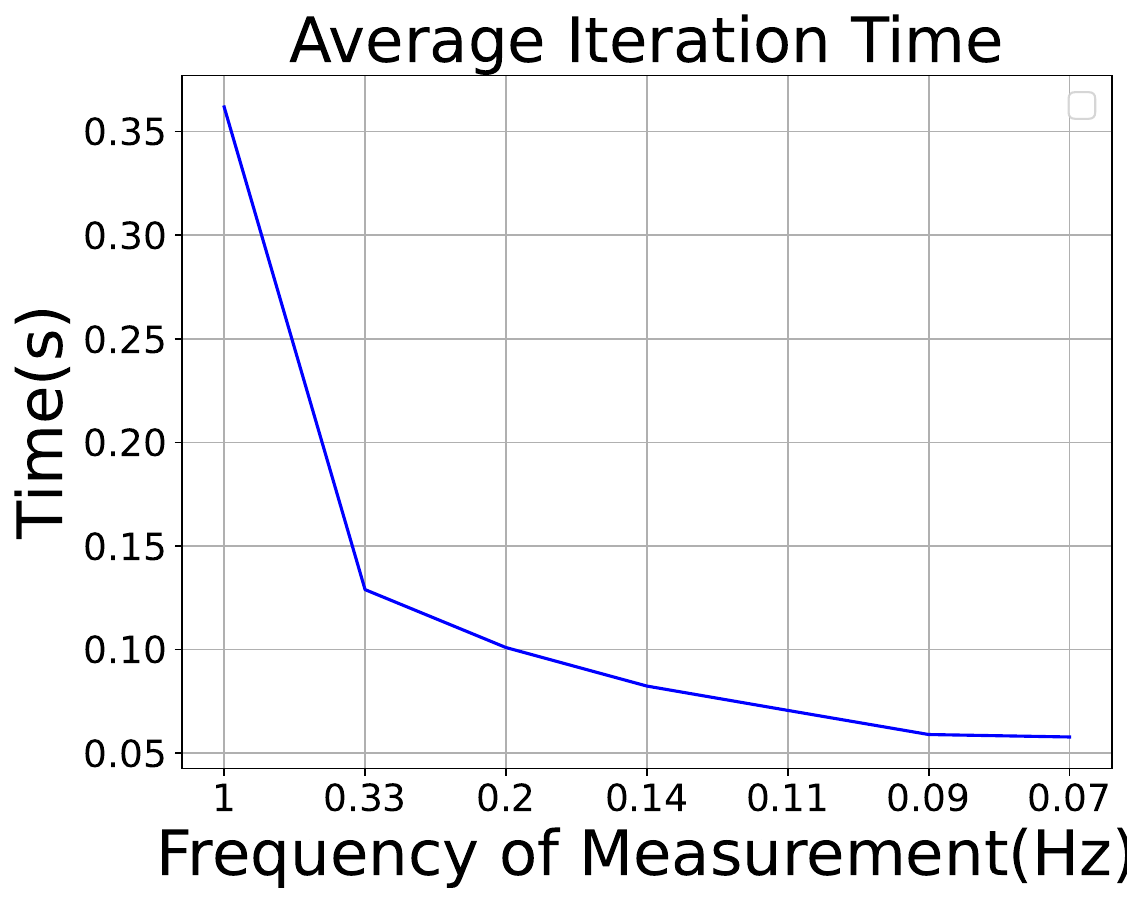}}
    \caption{\textcolor{black}{Average time of iteration for different frequency of measurement between pursuers and evader.} 
    }
    \label{freq_avg}
\end{figure}

\textcolor{black}{To demonstrate that the proposed method is able to solve the problem even when the measurement frequency is low, the method was evaluated at a measurement frequency of 0.2 Hz.
}
\begin{figure}[H]
    \centering
    \includegraphics[scale=0.3]{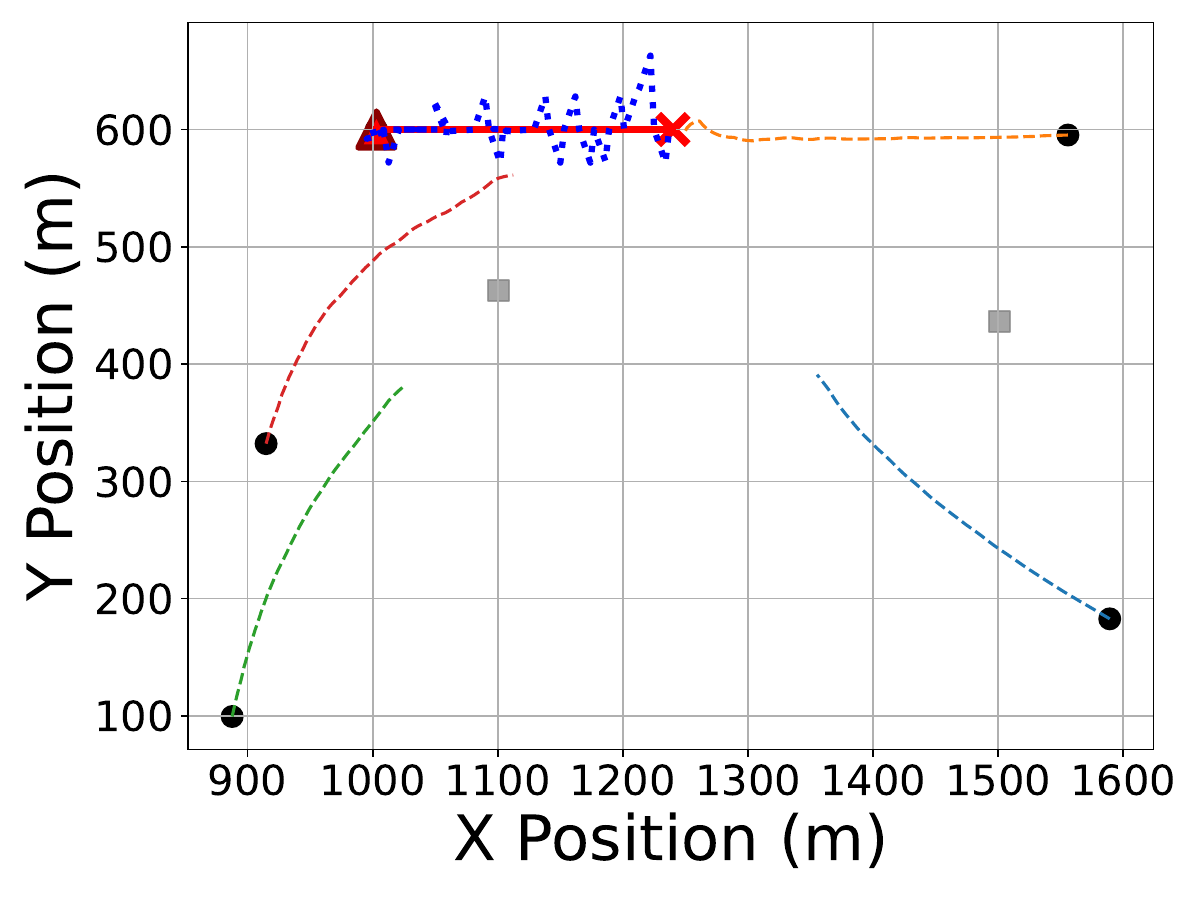}

    \vspace{0.5em} 

\newcommand{\legenditem}[2]{%
  \makebox[1.4cm][l]{#1} #2
}

\begin{tabular}{ll}
  \legenditem{\tikz[baseline=-0.6ex]\draw[line width=1pt, black, dashed] (0,0) -- (1.0cm,0);}{Pursuer 1} & 
  \legenditem{\tikz[baseline=-0.6ex]\draw[line width=1pt, red] (0,0) -- (1.0cm,0);}{Evader ground truth} \\

  \legenditem{\tikz[baseline=-0.6ex]\draw[line width=1pt, orange, dashed] (0,0) -- (1.0cm,0);}{Pursuer 2} & 
  \legenditem{\tikz[baseline=-0.6ex]\draw[line width=1pt, black, dotted] (0,0) -- (1.0cm,0);}{Evader estimate} \\

  \legenditem{\tikz[baseline=-0.6ex]\draw[line width=1pt, green, dashed] (0,0) -- (1.0cm,0);}{Pursuer 3} & 
  \legenditem{\tikz[baseline=-0.6ex]\filldraw[red] (0,0.06) -- (-0.05,-0.04) -- (0.05,-0.04) -- cycle;}{Evader start point} \\

  \legenditem{\tikz[baseline=-0.6ex]\draw[line width=1pt, red, dashed] (0,0) -- (1.0cm,0);}{Pursuer 4} & 
  \legenditem{\tikz[baseline=-0.6ex]\draw[red, thick] (-0.05,-0.05) -- (0.05,0.05) (-0.05,0.05) -- (0.05,-0.05);}{Evader final point} \\

  \legenditem{\tikz[baseline=-0.6ex]\filldraw[black] (0,0) circle (0.07);}{Pursuer start point} & 
  \legenditem{\tikz[baseline=-0.6ex]\filldraw[gray] (0,0) rectangle (0.12,0.12);}{Obstacles} \\
\end{tabular}

    \caption{\textcolor{black}{Performance when the frequency of measurement is 0.2 Hz.}}
    \label{freq}
\end{figure}
\textcolor{black}{This reduces computation time, as the frequency of measurement is decreased.
As shown in Fig.~\ref{freq_avg}, when the measurement frequency between pursuers and the evader decreases, the average time per iteration also drops.}
\textcolor{black}{Another experiment was conducted to show that, even when pursuers have different levels of odometry noise, they can still estimate the pose of the evader and plan to catch it.}
\begin{figure}[H]
    \centering
    \includegraphics[scale=0.3]{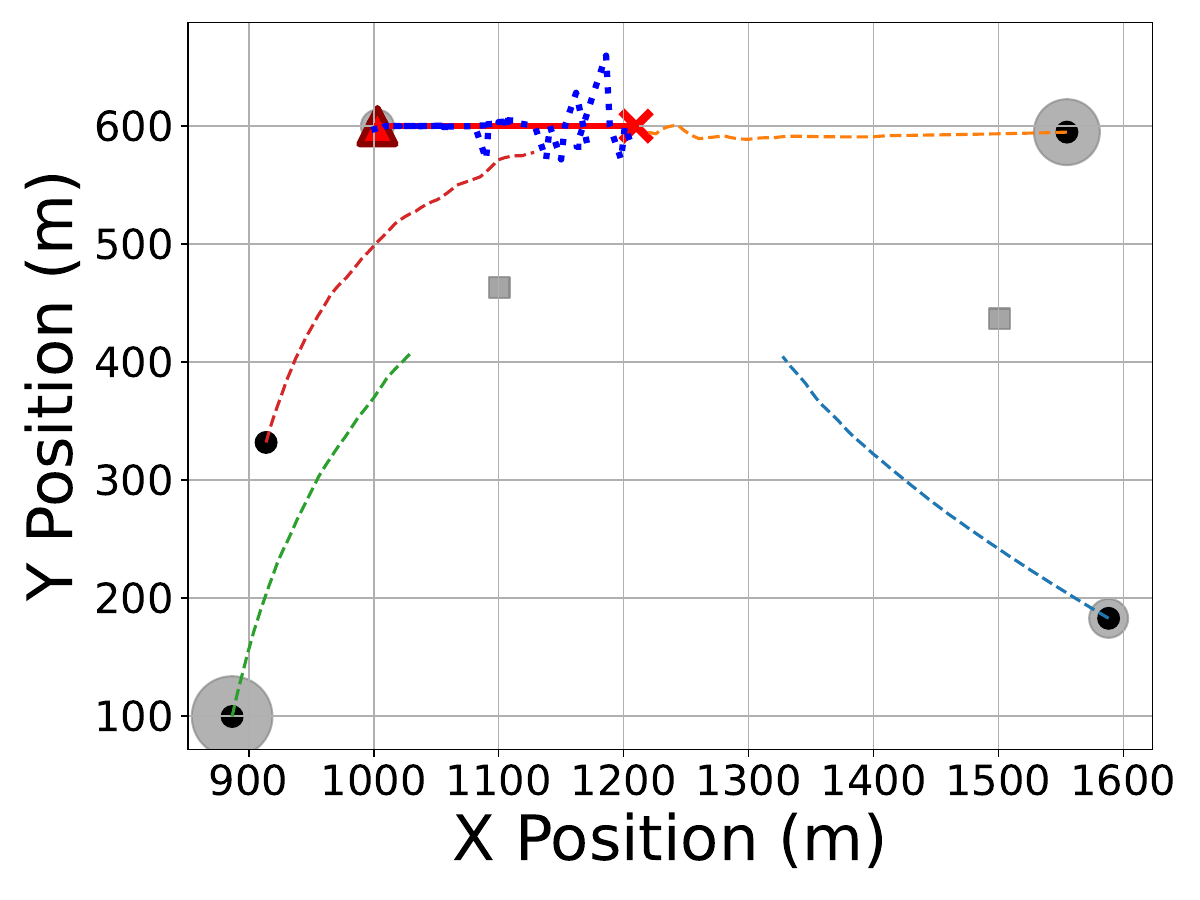}

    \vspace{0.5em} 

\newcommand{\legenditem}[2]{%
  \makebox[1.4cm][l]{#1} #2
}

\begin{tabular}{ll}
  \legenditem{\tikz[baseline=-0.6ex]\draw[line width=1pt, black, dashed] (0,0) -- (1.0cm,0);}{Pursuer 1} & 
  \legenditem{\tikz[baseline=-0.6ex]\draw[line width=1pt, red] (0,0) -- (1.0cm,0);}{Evader ground truth} \\

  \legenditem{\tikz[baseline=-0.6ex]\draw[line width=1pt, orange, dashed] (0,0) -- (1.0cm,0);}{Pursuer 2} & 
  \legenditem{\tikz[baseline=-0.6ex]\draw[line width=1pt, black, dotted] (0,0) -- (1.0cm,0);}{Evader estimate} \\

  \legenditem{\tikz[baseline=-0.6ex]\draw[line width=1pt, green, dashed] (0,0) -- (1.0cm,0);}{Pursuer 3} & 
  \legenditem{\tikz[baseline=-0.6ex]\filldraw[red] (0,0.06) -- (-0.05,-0.04) -- (0.05,-0.04) -- cycle;}{Evader start point} \\

  \legenditem{\tikz[baseline=-0.6ex]\draw[line width=1pt, red, dashed] (0,0) -- (1.0cm,0);}{Pursuer 4} & 
  \legenditem{\tikz[baseline=-0.6ex]\draw[red, thick] (-0.05,-0.05) -- (0.05,0.05) (-0.05,0.05) -- (0.05,-0.05);}{Evader final point} \\

  \legenditem{\tikz[baseline=-0.6ex]\filldraw[black] (0,0) circle (0.07);}{Pursuer start point} & 
  \legenditem{\tikz[baseline=-0.6ex]\filldraw[gray] (0,0) rectangle (0.12,0.12);}{Obstacles} \\
\end{tabular}

    \caption{\textcolor{black}{Performance when pursuers have different levels of odometry sensor noise. The shaded circle around each pursuer's start point indicates the level of noise, representing the noise covariance matrix.}}
    \label{differentnoise}
\end{figure}
\textcolor{black}{As shown in Fig.~\ref{differentnoise}, although the pursuers have different levels of odometry sensor noise, the proposed approach allows them to estimate the evader’s pose and plan to catch it. }
\subsubsection{Number of Robots}
\label{number_robots}
\begin{figure}[t!]
    \centering
    \begin{subfigure}[b]{0.49\linewidth}
        \includegraphics[width=\linewidth]{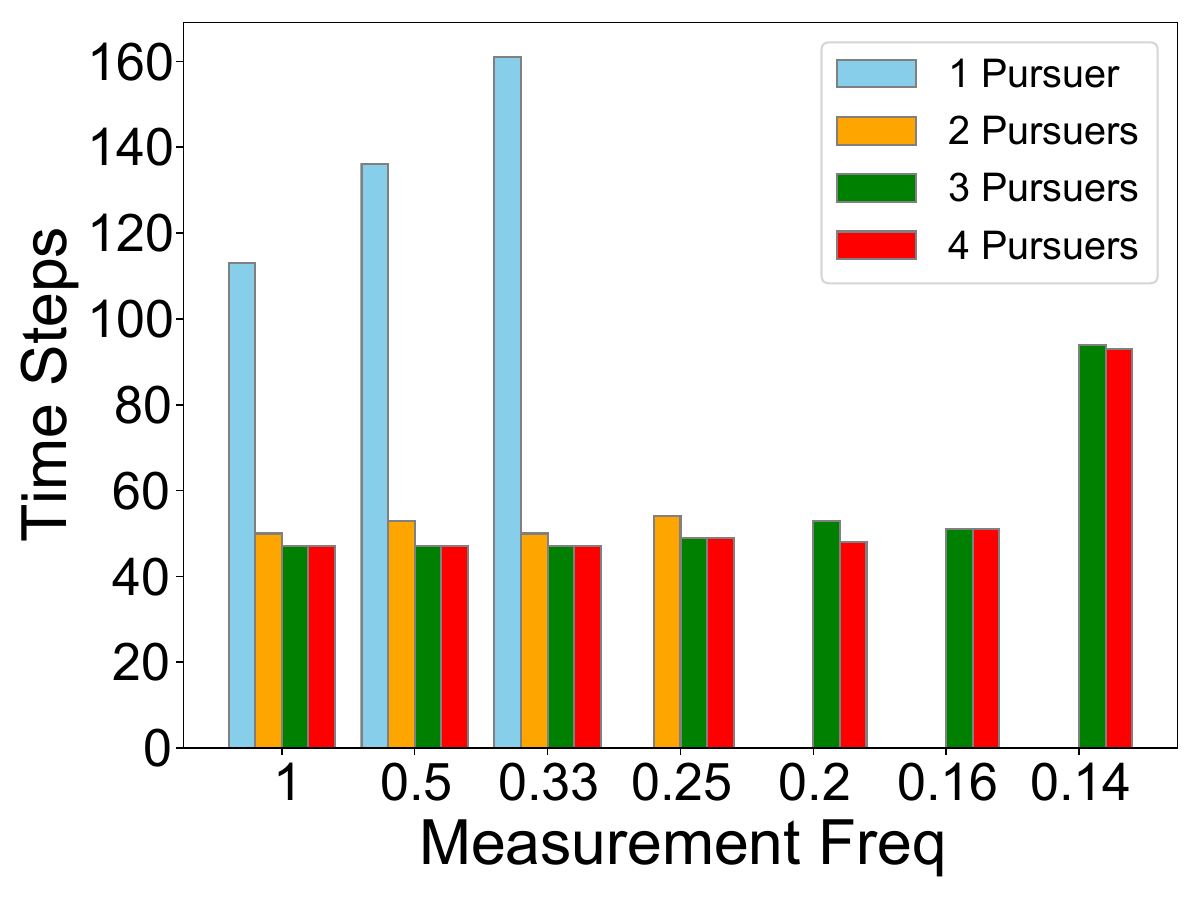}
        \label{fig:sub1}
        \vspace{-.5 cm}
    \end{subfigure}
    \begin{subfigure}[b]{0.49\linewidth}
        \includegraphics[width=\linewidth]{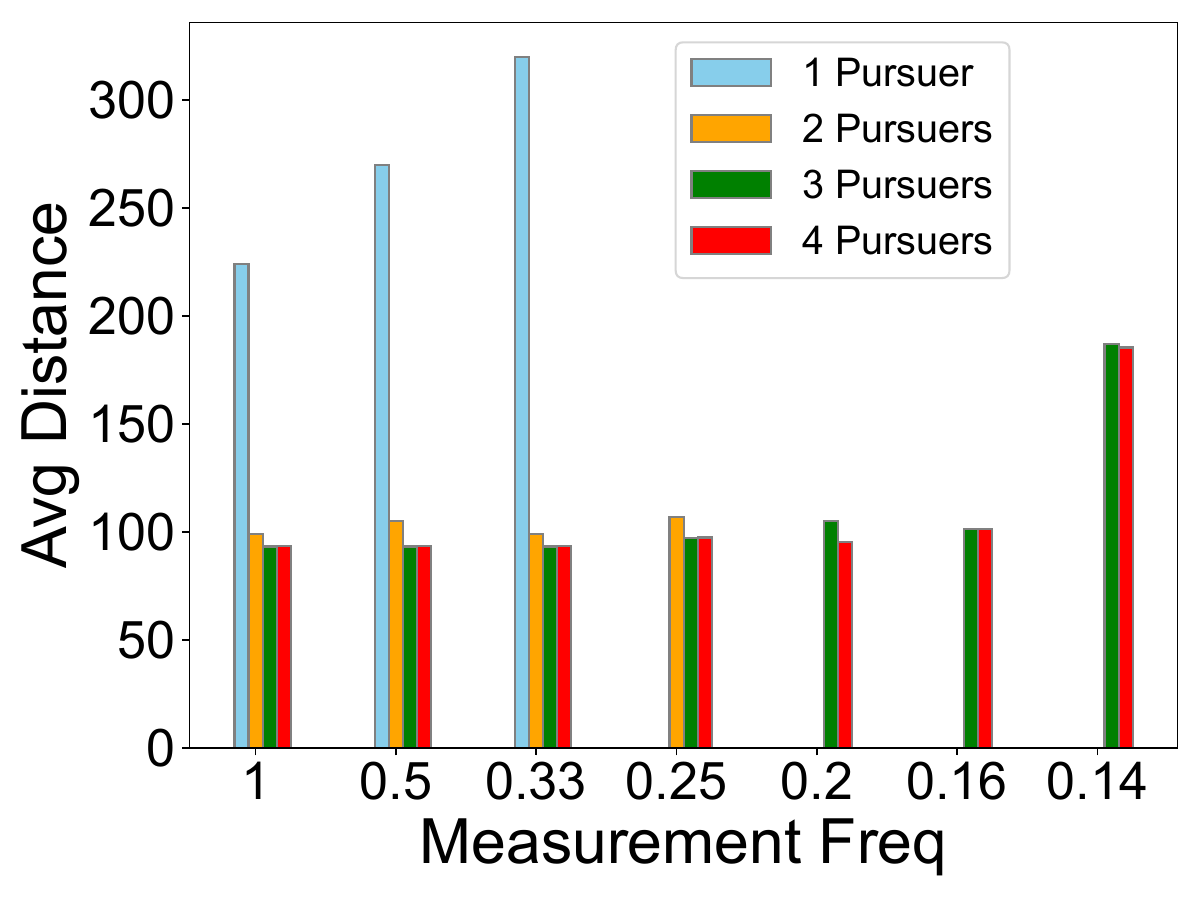}
        \label{fig:sub2}
        \vspace{-.5 cm}
    \end{subfigure}
    \vspace{2mm}
    \caption{Average distance traveled by pursuers and time to capture the evader comparison among different frequency of measurements and number of pursuers. } \label{fig:main_comparison3}
\end{figure}

\begin{figure}[t!]
    \centering
    \begin{subfigure}[b]{0.49\linewidth}
        \includegraphics[width=\linewidth]{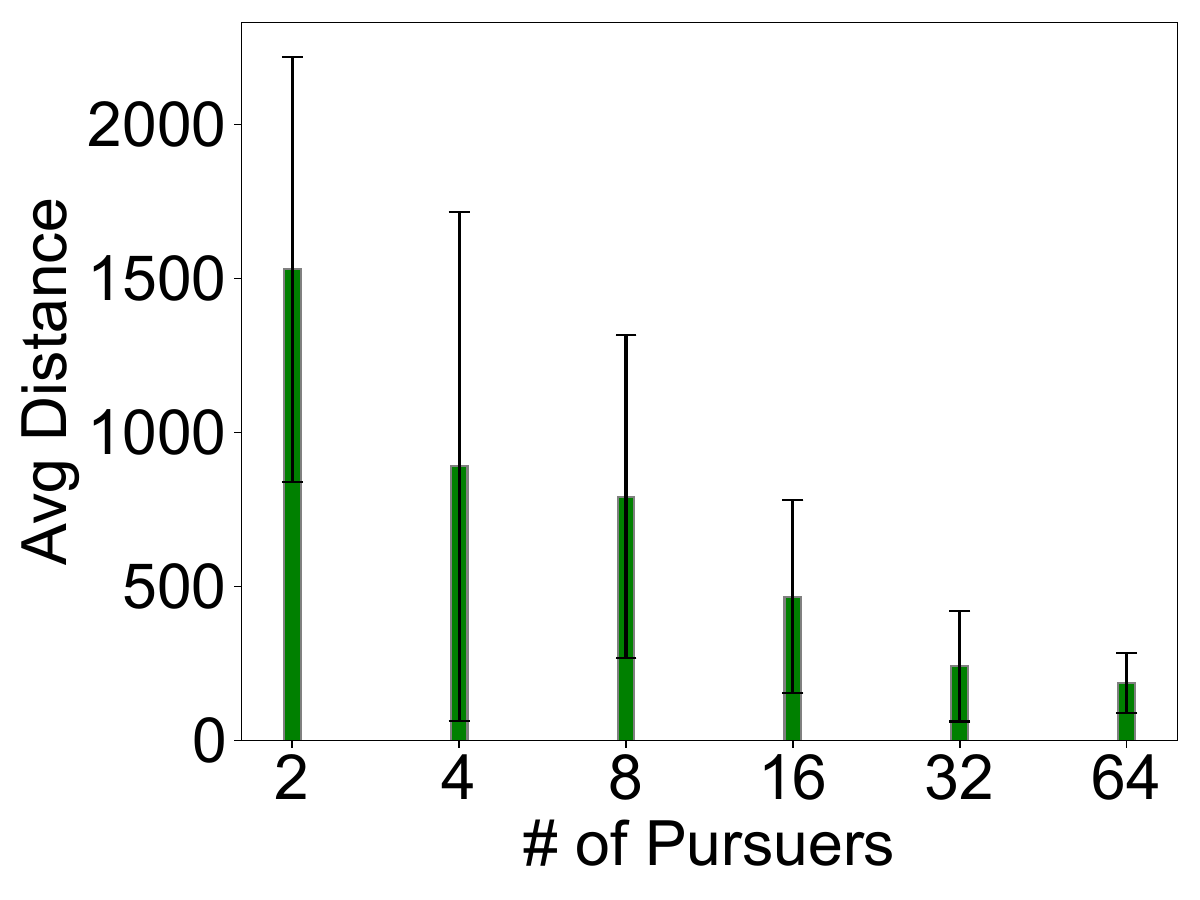}
        \label{fig:sub1}
        \vspace{-.5 cm}
    \end{subfigure}
    \begin{subfigure}[b]{0.49\linewidth}
        \includegraphics[width=\linewidth]{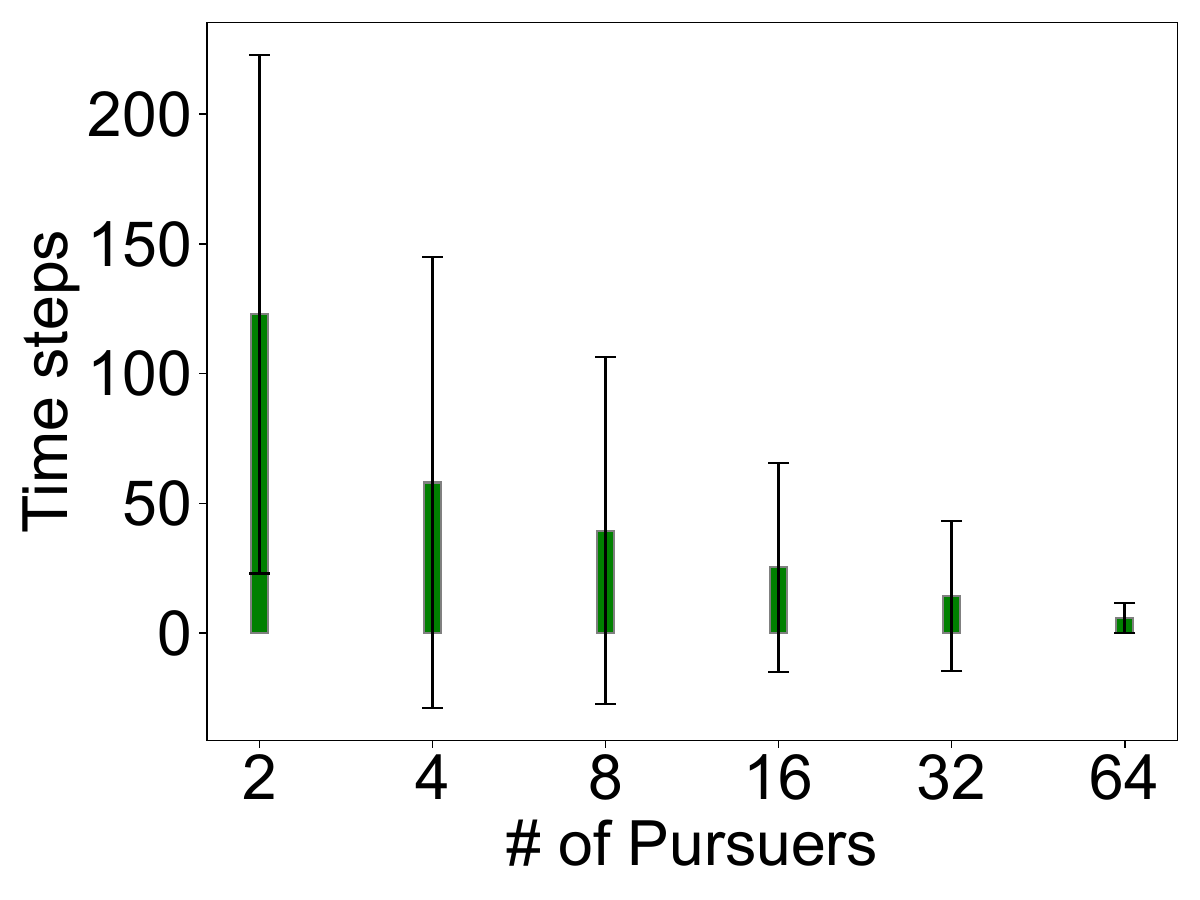}
        \label{fig:sub2}
        \vspace{-.5 cm}
    \end{subfigure}
    \vspace{-4mm}
    \caption{    Average distance traveled by pursuers (left) and time to capture the evader (right) for a different number of pursuers ($N_p$). 
    } \label{fig:main_comparison2_num_robots}
\end{figure}

The proposed method allows for flexible measurement frequency, so interactions between pursuers and the evader do not need to occur at every time step.
In Fig.~\ref{fig:main_comparison3}, the effect of the frequency of measurement when we have different numbers of pursuers is exhibited. 
Decreasing the frequency of measurement results in a longer time to capture the evader.
As can be seen in  Fig.~\ref{fig:main_comparison2_num_robots}, different number of robots have been used to evaluate the impact on evader capture time and the distance pursuer traveled. With each combination tested 10 times, 
it was found that increasing the number of pursuers reduces both the capture time and the average distance traveled.
\subsubsection{Uncertainty Ellipse}
\label{sec:uncertainty}
\begin{figure}[t!]
    \centering
    \begin{subfigure}[b]{0.49\linewidth}
        \includegraphics[width=\linewidth]{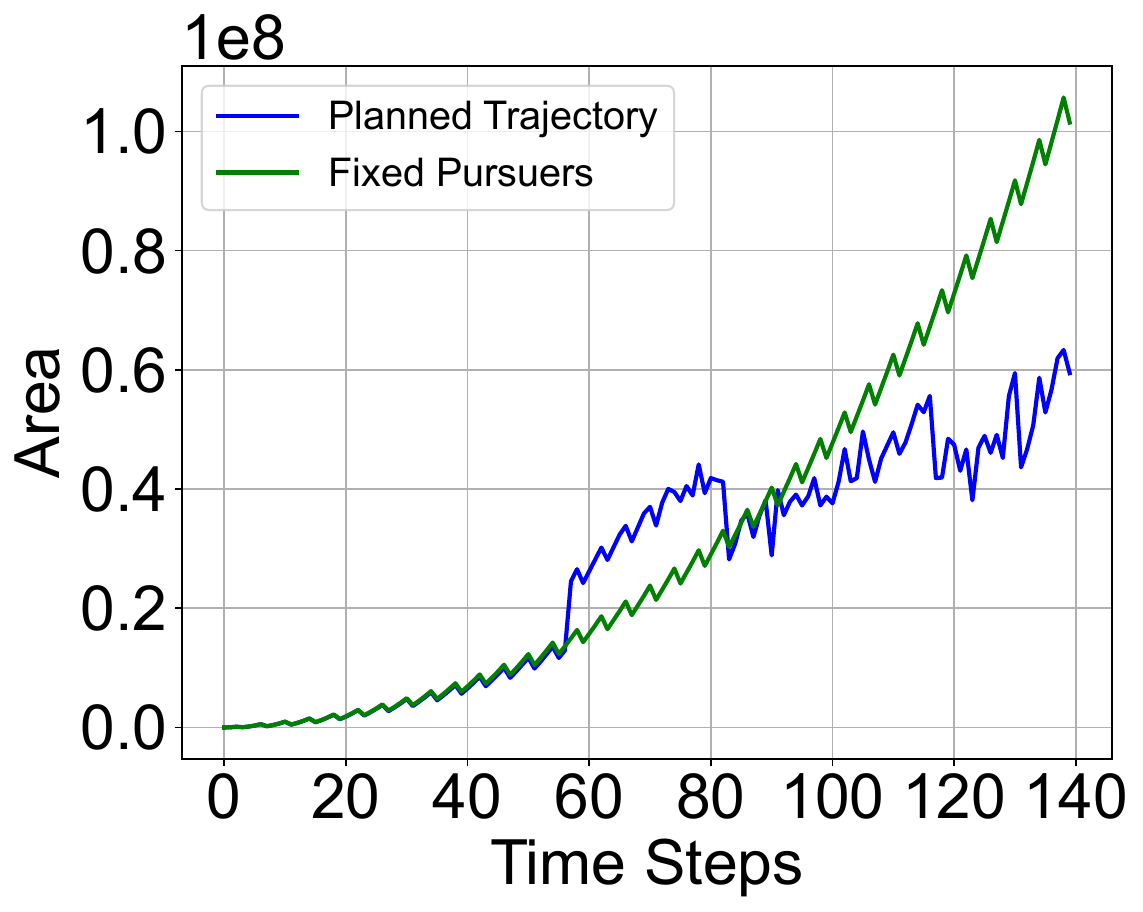}
        \label{fig:sub1}
        \vspace{-.5 cm}
    \end{subfigure}
    \begin{subfigure}[b]{0.49\linewidth}
        \includegraphics[width=\linewidth]{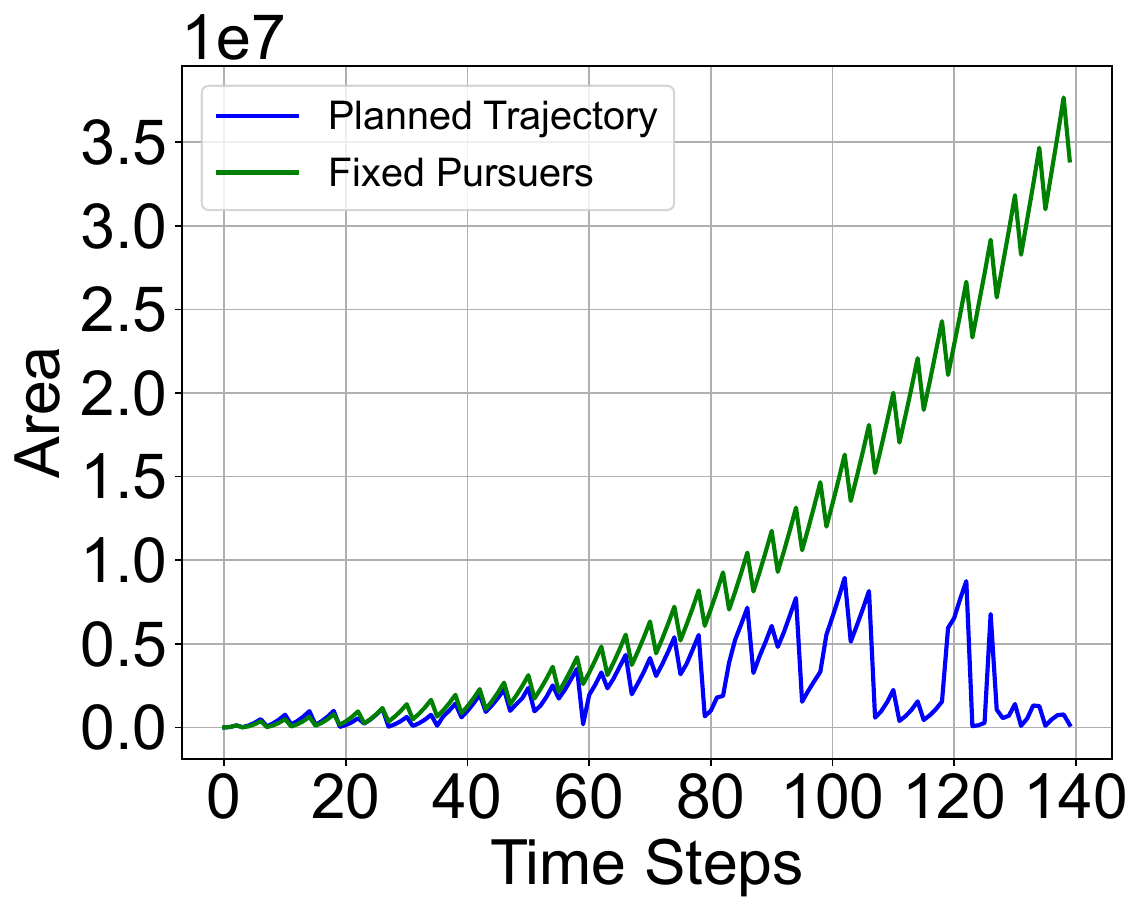}
        \label{fig:sub2}
        \vspace{-.5 cm}
    \end{subfigure}
    \vspace{-4mm}
    \caption{The comparison of the uncertainty ellipse area for the scenarios where the pursuers are planned and fixed, with no landmark in the scene (left), and with one landmark in the scene (right).
    } \label{fig:ellipse_compare}
\end{figure}

In this subsection, the effectiveness of the robot's planning for catching the evader and decreasing the uncertainty is investigated.
\begin{figure}[ht!]
    \centerline{\includegraphics[scale=0.35]{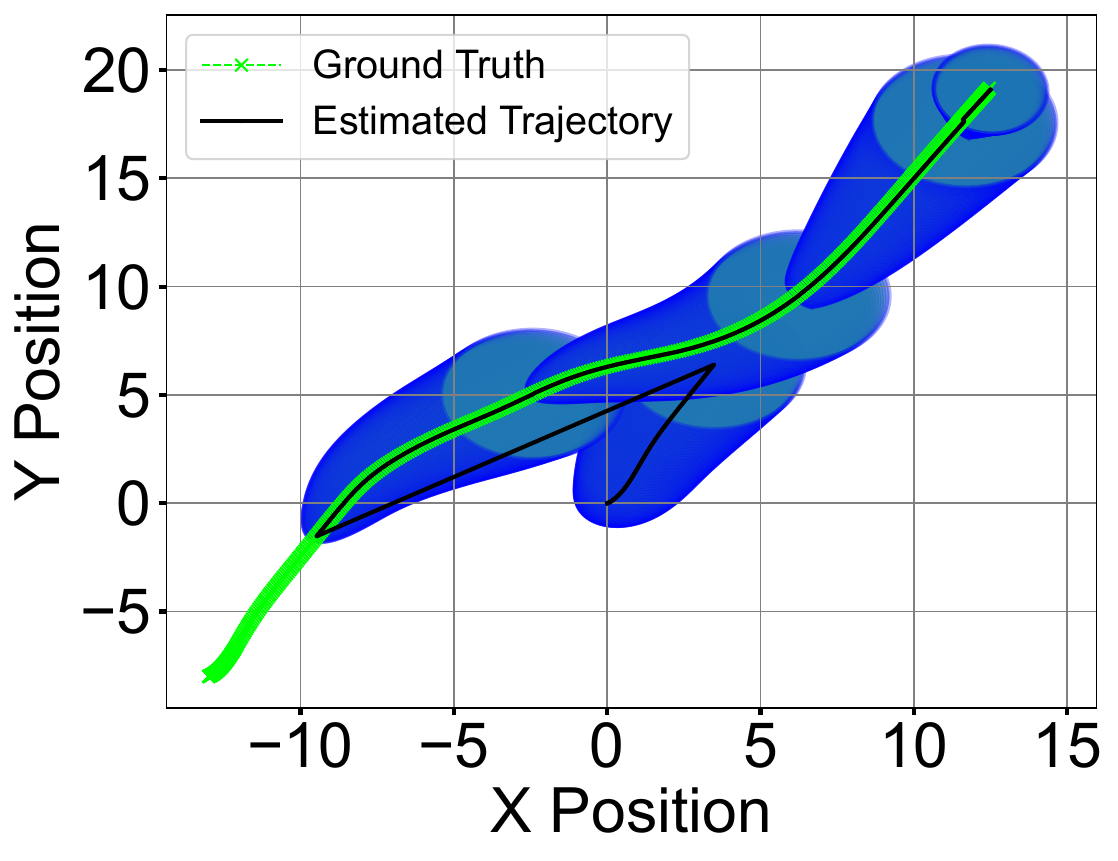}}
    \caption{The uncertainty minimization of the estimated position. The green line shows the ground truth and the black line shows the estimated position. The frequency of measurement from pursuers to evader is 0.01 Hz and the blue ellipse shows the uncertainty ellipse that is scaled by 1/1000. 
    }
    \label{ellipse}
\end{figure}
Fig.~\ref{ellipse} illustrates that pursuers reduce uncertainty by adjusting their movement. \textcolor{black}{During the experiment and the planning of the pursuers, when a measurement is taken from the pursuers of the evader, the uncertainty decreases. While other approaches lack consideration of uncertainty in their predictions, the proposed method accounts for it and also minimizes it. \textcolor{black}{In Fig. \ref{ellipse_area_freq}, the area of the uncertainty ellipse for different measurement frequencies is shown. As can be seen, decreasing the measurement frequency results in an increase in the area of uncertainty. The area is shown in log scale in Fig \ref{ellipse_area_freq}.}\\
\begin{figure}[ht!]
    \centerline{\includegraphics[scale=0.35]{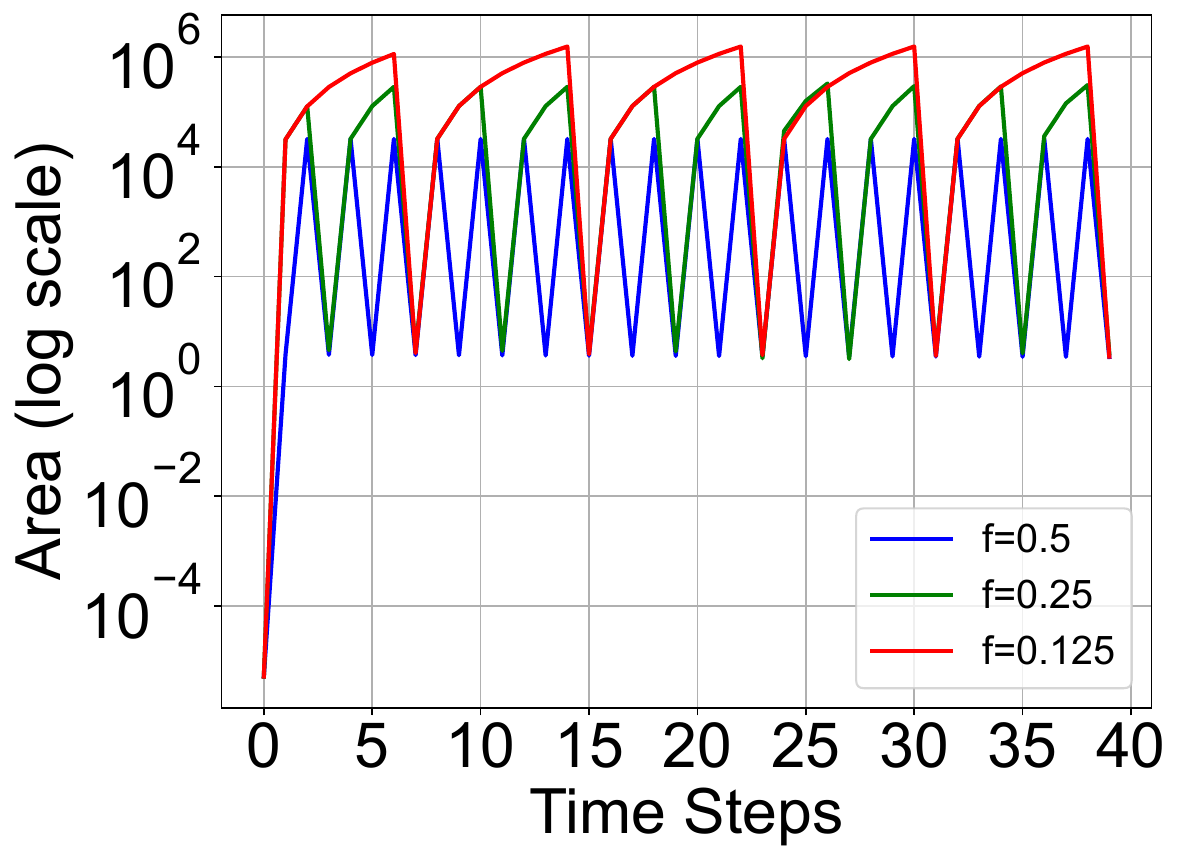}}
    \caption{The area of the uncertainty ellipse for different frequencies of measurements arriving from the pursuers.  
    }
    \label{ellipse_area_freq}
\end{figure}
}
\textcolor{black}{In Fig. \ref{fig:ellipse_compare}, the effectiveness of the proposed planning method is illustrated in environments with either no obstacles or a limited number of obstacles. As shown, when there are no obstacles (left plot), the area of the uncertainty ellipse is smaller compared to the scenario where the pursuers remain $\text{stationary}$. In the case where one obstacle is present (right plot), the reduction in the area of the uncertainty ellipse becomes even more pronounced.}
\subsubsection{Dropped Messages}

\begin{figure}[ht!]
    \centerline{\includegraphics[scale=0.35]{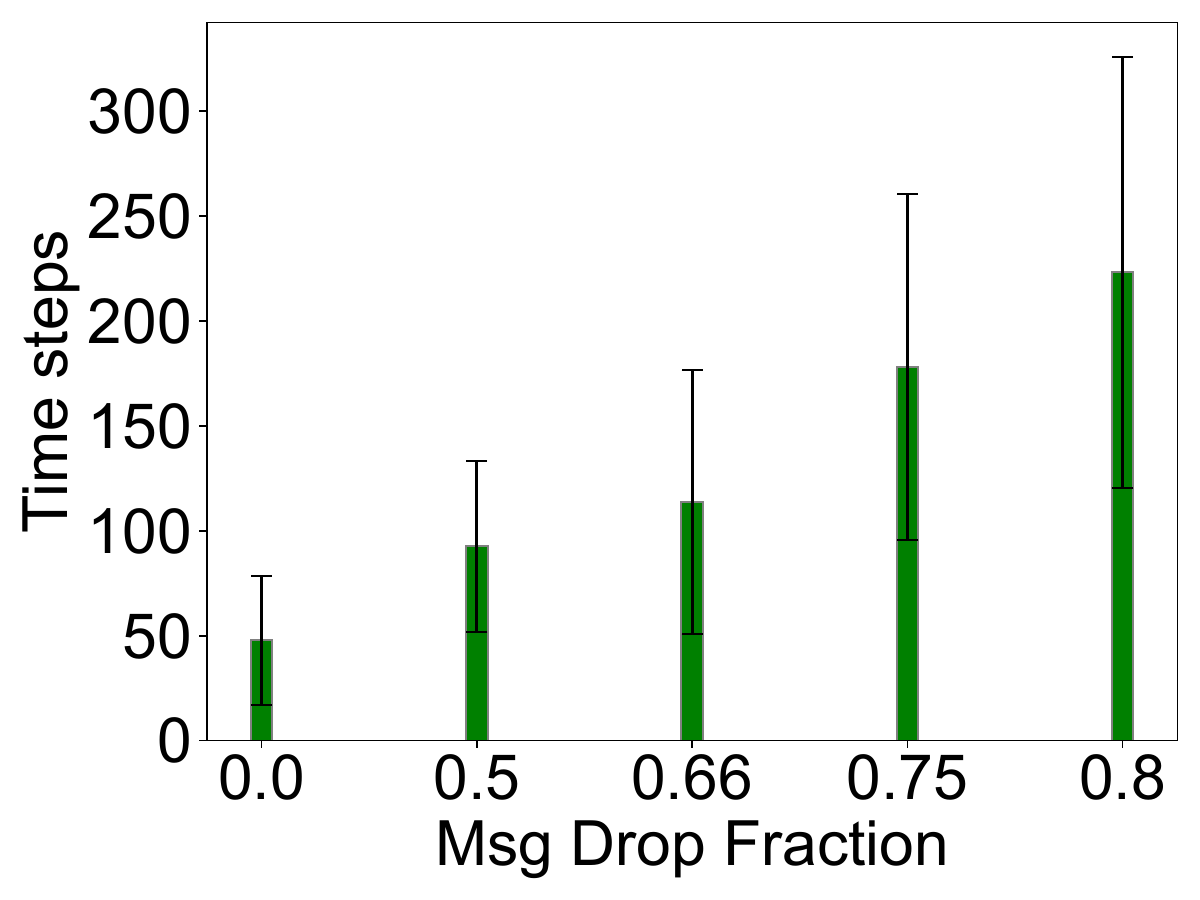}}
    \caption{The average time to catch the evader when different fraction of messages are dropped.
    }
    \label{fig:dropped_msgs}
\end{figure}

In this experiment, four pursuers are placed randomly in the simulation environment for 30 trials at each fraction of dropped messages. As shown in Fig. \ref{fig:dropped_msgs}, as the frequency of dropped messages increases, the average time step required to catch the evader also increases. Additionally, these results demonstrate the reliability of the proposed method in handling dropped $\text{messages}$.

\subsection{Real-world Experiment}
\textcolor{black}{
\textcolor{black}{To demonstrate that the provided solution is applicable in real-world scenarios, an experiment was designed to test it in practical situations.}
For this experiment, TurtleBots were utilized, and the development was conducted in ROS2. \textcolor{black}{The experiment shows the performance of the proposed solution even in real-world scenarios.} Two pursuers and one evader were considered. The localization of the robots was based on the odometry of each robot, which increases the uncertainty of localization and prediction.} 
\begin{figure}[t!]
    \centering

    \begin{subfigure}[b]{0.49\linewidth}
        \includegraphics[width=\linewidth]{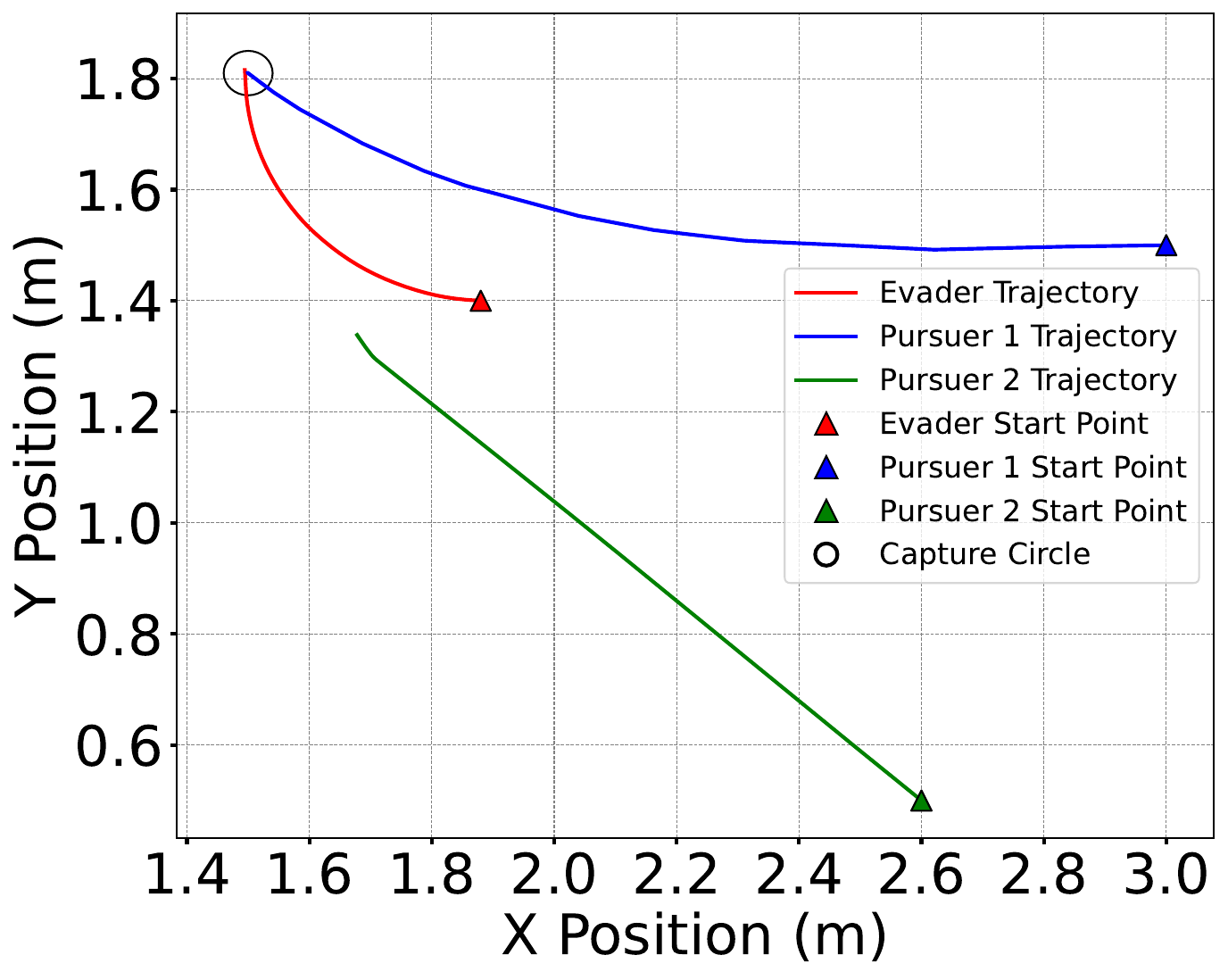}
        \label{fig:sub1}
        \vspace{-.5 cm}
    \end{subfigure}
    \begin{subfigure}[b]{0.49\linewidth}
        \includegraphics[width=\linewidth]{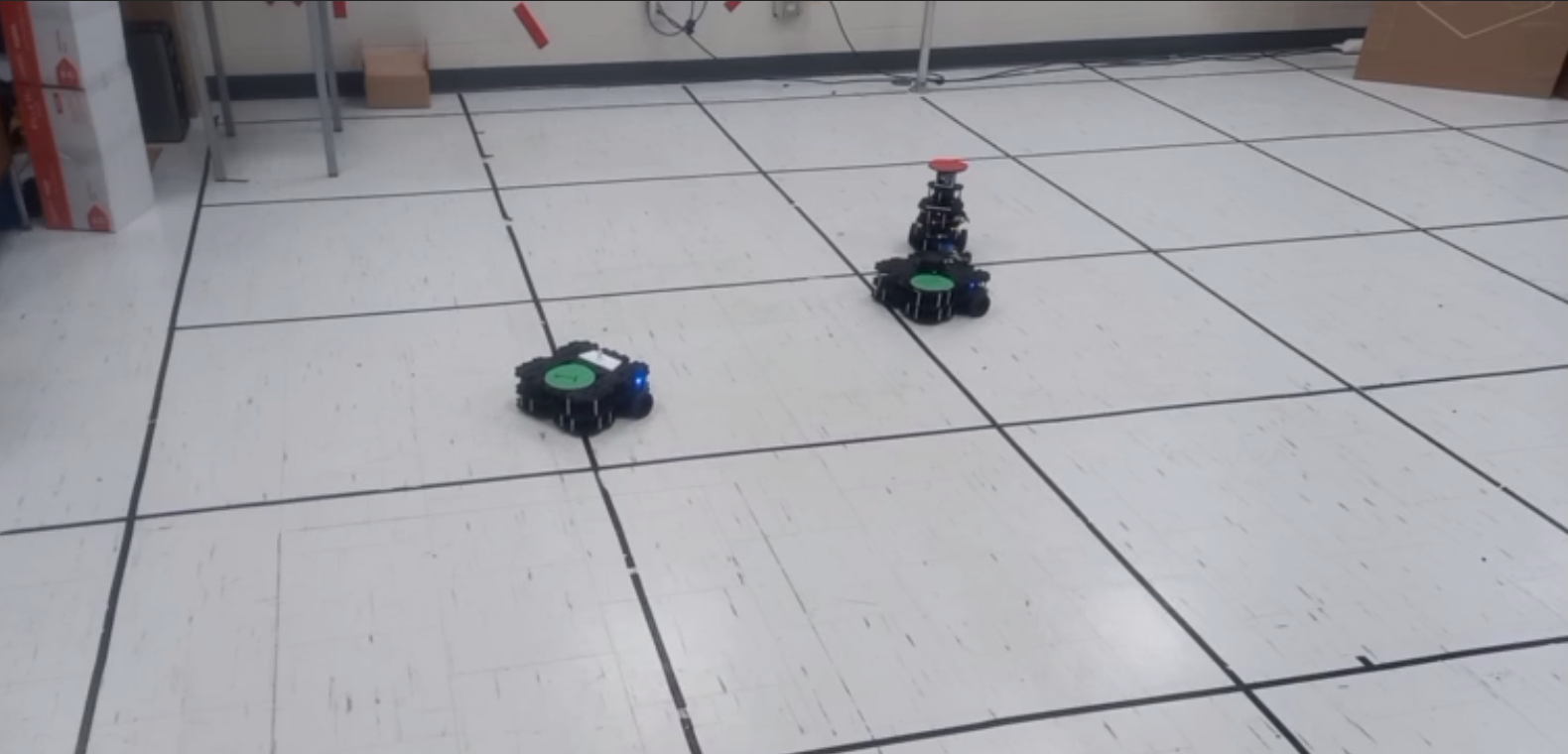}
        \label{fig:sub2}
        \vspace{0.5 cm}
    \end{subfigure}
    \vspace{-4mm}
    \caption{\textcolor{black}{The ground-truth trajectory of the evader is represented in red. The positions of the pursuers are shown in blue and green. The initial position of pursuers is marked by triangles. The dashed circle shows the catching criteria.} 
    } \label{real_world}
\end{figure}
\textcolor{black}{As illustrated in Fig.~\ref{real_world}, although the pursuers are far from the evader.  The speed of the pursuers in this experiment is twice the speed of the evader ($V_p = 2V_q$).} Videos of the experiments are available on the project website:~\href{https://sites.google.com/view/pursuit-evasion}{https://sites.google.com/view/pursuit-evasion}

\textcolor{ black}{\subsection{Discussion}}
\textcolor{ black}{Utilizing factor graphs, our method (FG-PE) provides a unified framework that seamlessly combines estimation and planning, allowing pursuers to track the evader’s motion while choosing an efficient capture path. The advantages of FG-PE are rooted in the core properties of factor graphs and their robust architecture. The main reasons for the superior performance of FG-PE are:}

\noindent \textcolor{ black}{1) FG-PE Explicitly Accounts for Uncertainty: The probabilistic nature of factor graphs allows FG-PE to directly model and minimize uncertainty in predictions. When a new measurement is introduced, the factor graph's Jacobian matrix is updated to incorporate this richer information, leading to more accurate estimations and reduced uncertainty. As summarized in Sec.~\ref{sec:uncertainty} and Fig.~\ref{ellipse_area_freq}, increasing the frequency of measurement reduces the capture time by 16\%. Moreover, the experiments show that for different measurement frequencies, the proposed solution can decrease the uncertainty area by a factor of $10^5$ on average.
}

\noindent \textcolor{ black}{2) FG-PE Handles Noise and Sparse Data: Factor graphs are highly effective at handling sparse and noisy measurements. This allows FG-PE to achieve accurate trajectory estimations and strategic planning even with fewer measurements, resulting in lower memory usage and computational overhead. Fig.~\ref{differentnoise} shows that, despite the presence of different odometry noise levels, which result in poorer estimates of the evader's position, FG-PE is still able to generate effective plans that enable the pursuers to capture the evader. }

\noindent \textcolor{ black}{3) FG-PE Provides a Flexible and Scalable Framework: The factor graph's design allows it to adapt to varying numbers of pursuers and obstacles, making it easily scalable for multi-agent problems. As shown in Fig.~\ref{fig:main_comparison2_num_robots}, doubling the number of pursuers enables them to capture the evader faster, reducing the capture time by 55\%.} 

\noindent \textcolor{ black}{4) FG-PE Maintains Robustness: The method is robust and performs well even when some communication messages are dropped, as the graph structure can manage these lost connections. As shown in Fig.~\ref{fig:dropped_msgs}, even when the fraction of dropped messages increases, the pursuers are still able to catch the evader. Doubling the fraction of dropped messages increases the capture time by 48\%.}

\textcolor{ black}{Overall, the ability of FG-PE to accurately estimate trajectories and plan strategically within a single and unified framework explains its consistent outperformance, especially in challenging speed scenarios.}\\

\section{Conclusions}\label{sec:con}
In this paper, we propose a factor graph-based approach to solve the pursuit-evasion problem, which is capable of both tracking and planning within dynamic environments. The pursuit-evasion problem is a critical challenge in robotics and multi-agent systems, where multiple pursuers must effectively track and capture an evader. Our proposed approach is robust against dropped messages, which is essential for real-time applications where communication may be unreliable. Additionally, it is designed to be scalable to accommodate different numbers of robots and obstacles, allowing it to be applicable in various operational contexts.

To reduce optimization time, we utilize a single optimizer, which enhances scalability by reducing dimensionality and computational overhead. This design choice ensures that the algorithm remains efficient, even as the complexity of the environment increases. Furthermore, we incorporate additional factors into the factor graph to add more constraints, which enhances the accuracy of the estimation and tracking processes. This flexibility allows the system to adapt to changing conditions and maintain performance under \mbox{various }scenarios.

The proposed method is rigorously compared with state-of-the-art techniques, demonstrating a significant reduction in both the time required to capture the evader and the average distance traveled by the pursuers. The conducted hardware experiments confirm the robustness of our approach, showcasing its effectiveness in real-world applications. Furthermore, scalability tests reveal that computation time decreases even as environmental complexity increases, indicating that our method can handle more demanding situations without sacrificing performance.

In the future, we aim to integrate Gaussian processes into the solution to enable distributed message passing, which will further decrease the time needed to capture the evader. This integration could enhance the collaborative capabilities of the robots, allowing them to share information more efficiently and respond dynamically to changes in the environment. Another potential avenue for future work could involve solving the pursuit-evasion problem in the presence of dynamic obstacles and multiple evaders, which would significantly add complexity and realism to the pursuit-evasion scenario. By addressing these challenges, we aim to further improve the applicability of our approach in real-world situations where conditions are unpredictable and ever-changing.

 \bibliographystyle{elsarticle-num} 
 \bibliography{doc}





\end{document}